\documentclass[journal,compsoc]{IEEEtran}
\ifCLASSOPTIONcompsoc
  \usepackage[nocompress]{cite}
\else
  \usepackage{cite}
\fi

\ifCLASSINFOpdf
\else
\fi

\usepackage{xcolor}
\usepackage{color}
\definecolor{mygray}{RGB}{193,203,215}
\usepackage{booktabs}
\usepackage{multirow}
\usepackage{colortbl}
\usepackage{graphicx}
\usepackage{caption}
\DeclareCaptionFont{blue}{\color{blue}}

\usepackage{array}
\usepackage{dblfloatfix}
\usepackage{balance}
\usepackage{enumitem}
\usepackage{url}
\usepackage{dingbat}
\newcolumntype{C}[1]{>{\centering\arraybackslash}m{#1}}

\usepackage[framemethod=tikz]{mdframed}
\definecolor{mycolor}{rgb}{0.9, 0.0, 0.0}
\newmdenv[innerlinewidth=0.5pt, roundcorner=4pt,linecolor=mycolor,innerleftmargin=6pt,
innerrightmargin=6pt,innertopmargin=6pt,innerbottommargin=6pt]{rolebox}
\definecolor{mycolor1}{rgb}{0.0, 0.0, 0.9}
\newmdenv[innerlinewidth=0.5pt, roundcorner=4pt,linecolor=mycolor1,innerleftmargin=6pt,
innerrightmargin=6pt,innertopmargin=6pt,innerbottommargin=6pt]{commentbox}

\usepackage{bigstrut}
\usepackage{multirow}
\usepackage{amsmath}
\usepackage{subfigure}

\makeatletter
\long\def\@IEEEtitleabstractindextextbox#1{\parbox{0.922\textwidth}{#1}}
\makeatother

\usepackage[rawfloats=true]{floatrow} 
\restylefloat{figure} 
\restylefloat{table} 

\usepackage{cite}
\usepackage{amsmath,amssymb,amsfonts}
\usepackage{algorithmic}
\usepackage{caption,graphicx}
\usepackage{textcomp}
\usepackage{xcolor}
\usepackage{mathrsfs}
\usepackage{pgfplots}

\usepackage{amssymb}
\usepackage{pifont}
\usepackage{url}
\usepackage{textcomp}
\usepackage{xcolor}
\usepackage{mathrsfs}
\usepackage{booktabs} 
\usepackage{tikz}
\usepackage{dblfloatfix}
\usepackage{bigstrut}
\usepackage{multirow}
\usepackage{color}
\usepackage{fancyhdr}
\usepackage{listings} 
\usepackage{changepage}
\usepackage{lipsum}
\usepackage{floatrow}
\usepackage{balance}
\usepackage{enumitem}
\usepackage{comment}
\usepackage{breqn}
\usepackage{dblfloatfix}
\usepackage{booktabs} %
\def\BibTeX{{\rm B\kern-.05em{\sc i\kern-.025em b}\kern-.08em
    T\kern-.1667em\lower.7ex\hbox{E}\kern-.125emX}}

\makeatletter
\long\def\@IEEEtitleabstractindextextbox#1{\parbox{0.922\textwidth}{#1}}
\makeatother

\begin{document}
\bstctlcite{IEEEexample:BSTcontrol}

%
\title{MIPGAN - Generating Strong and 
    High Quality Morphing Attacks Using Identity Prior Driven GAN}
%
%
%
\author{Haoyu Zhang$^\dag$, Sushma Venkatesh$^\dag$, Raghavendra Ramachandra$^\dag$ \\ Kiran Raja$^\dag$, Naser Damer$^\ddagger$,  Christoph Busch$^\dag$ \\
$^\dag$ Norwegian University of Science and Technology (NTNU), Norway\\  
	$^\ddagger$Fraunhofer Institute for Computer Graphics Research IGD, Darmstadt, Germany.\\
	\{\tt\small sushma.venkatesh; haoyu.zhang; raghavendra.ramachandra;kiran.raja;christoph.busch\} @ntnu.no\\
	\{\tt\small naser.damer\}@igd.fraunhofer.de\\
}
%
%

\markboth{Journal of \LaTeX\ Class Files,~Vol.~14, No.~8, August~2015}%
{Shell \MakeLowercase{\textit{et al.}}: Bare Demo of IEEEtran.cls for IEEE Journals}
%




\IEEEtitleabstractindextext{
\begin{abstract}
Face morphing attacks target to circumvent Face Recognition Systems (FRS) by employing face images derived from multiple data subjects (e.g., accomplices and malicious actors). Morphed images can be verified against contributing data subjects with a reasonable success rate, given they have a high degree of facial resemblance. 
The success of morphing attacks is directly dependent on the quality of the generated morph images. We present a new approach for generating strong attacks extending our earlier framework for generating face morphs. We present a new approach  using an Identity Prior Driven Generative Adversarial Network, which we refer to as \textit{MIPGAN (Morphing through Identity Prior driven GAN)}. The proposed MIPGAN is derived from the StyleGAN with a newly formulated loss function exploiting perceptual quality and identity factor to generate a high quality morphed facial image with minimal artefacts and with high resolution. We demonstrate the proposed approach's applicability to generate strong morphing attacks by evaluating its vulnerability against both commercial and deep learning based Face Recognition System (FRS) and demonstrate the success rate of attacks. Extensive experiments are carried out to assess the FRS's vulnerability against the proposed morphed face generation technique on three types of data such as digital images, re-digitized (printed and scanned) images, and compressed images after re-digitization from newly generated \textit{MIPGAN Face Morph Dataset}. The obtained results demonstrate that the proposed approach of morph generation poses a high threat to FRS. 
\end{abstract}

}

%
\IEEEpeerreviewmaketitle

\maketitle

\begin{figure}[b]
\parbox{\hsize}{\em
Haoyu Zhang, Sushma Venkatesh and Raghavendra Ramachandra contributed equally.
}\end{figure}

\section{Introduction}

\begin{figure*}[htp]
\begin{center}
    \centering
   \includegraphics[width=0.7\linewidth]{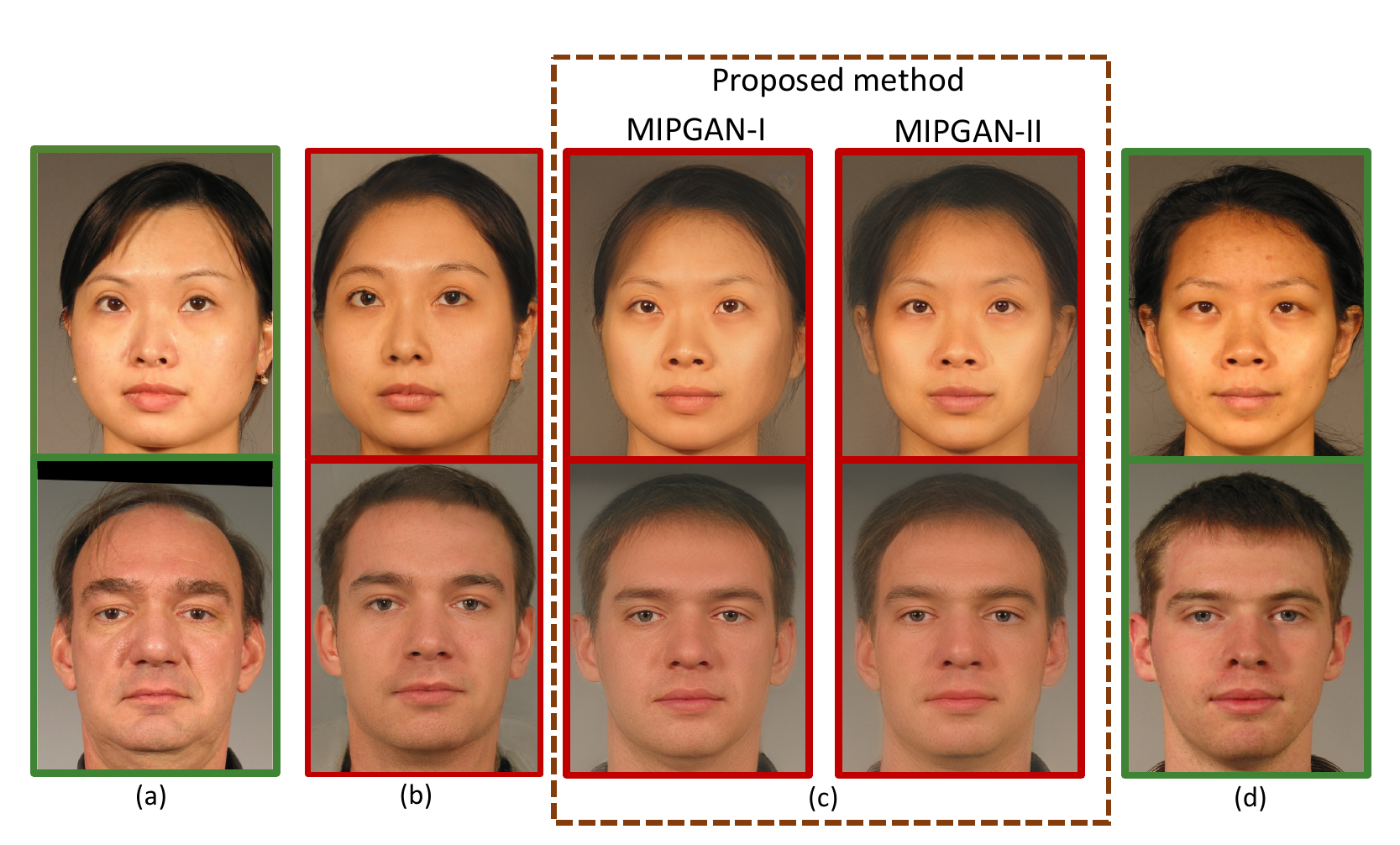}
	\captionof{figure}{Results from StyleGAN based face morphing \cite{MorphStyleGAN2020} and the proposed MIPGAN (a) Contributing subject 1  (b)  StyleGAN\cite{MorphStyleGAN2020} (c) Proposed method (d)  Contributing subject 2}
	\label{fig:iwbf2020-stylegan-morph-image}
\end{center}%
\end{figure*}
Face Recognition Systems (FRS) have provided ubiquitous ways of verifying an identity claim in many applications. FRS have been used in everyday applications from low-security applications such as smartphone unlocking to high-security applications such as identity verification in border control processes. Each of the applications mandate a chosen way of enrolment to FRS where either a supervised enrolment is carried out (for instance in on-boarding at bank premises) or unsupervised enrolment is requested (on-boarding for banking applications from home). While it provides a high degree of flexibility and convenience to users to initiate an enrolment process in an unsupervised manner, this potentially leads to a security risk: Without supervision, a data subject enrolling into the FRS can submit a face image which is manipulated, a printed face image, an image displayed from an electronic screen (e.g., iPad) or a silicone latex face mask \cite{ramachandra2017presentation}. In order to mitigate such attacks at the enrolment level, it is therefore essential to have a robust attack detection mechanism. While a number of works in recent years have been proposed on both conducting such attacks and detecting the attacks in a robust manner for printed attacks, display attacks and mask attacks, in this work we focus on a new kind of attack referred popularly as \textit{Morphing Attack}.

Face morphing is the process of combining two or more face images to generate a single face image that can resemble visually to all the contributing face images to a greater degree \cite{ferrara2014magic}. A good quality morphed face image is also effective in verifying against all contributing subjects by obtaining a comparison score that exceeds the pre-determined threshold (i.e., passes through FRS) \cite{ferrara2014magic,raghavendra2017transferable,raghavendra2016detecting, sotamdpaper}. While morphing can be conducted using multiple face images of different subjects, the effectiveness of morphed images is reported when the face images of similar ethnicity, gender and age group are considered \cite{raghavendra2017face,sotamdpaper,scherhag2017biometric}. This is primarily due to the fact that a morphed image should not only defeat the FRS but should also provide high visual similarity, in order to convince a human expert in a visual comparison process.

Face morphing attacks threaten FRS due to the current practices in the ID-document application process, where the biometric enrolment is carried out in an unsupervised manner in many countries. Countries like the UK and New Zealand allow citizens to upload a digital face image for various applications such as passport renewal \cite{ukpassport2019portal} and visa application \cite{nzvisa2019portal}. The capture process for such images is unsupervised. In a similar manner, many Asian countries 
and European countries (e.g. in The Netherlands \cite{netherlands2019portal}) request the applicant to submit a scanned face image for passport/visa/identity-card applications.  Given that the images are captured and submitted in an unsupervised setting, the applicant has vast opportunities to upload a morphed image with malicious intent underlining the need for robust Morphing Attack Detection (MAD) mechanisms.

\begin{figure*}[htp]
	\centering
	\includegraphics[width=0.9\linewidth]{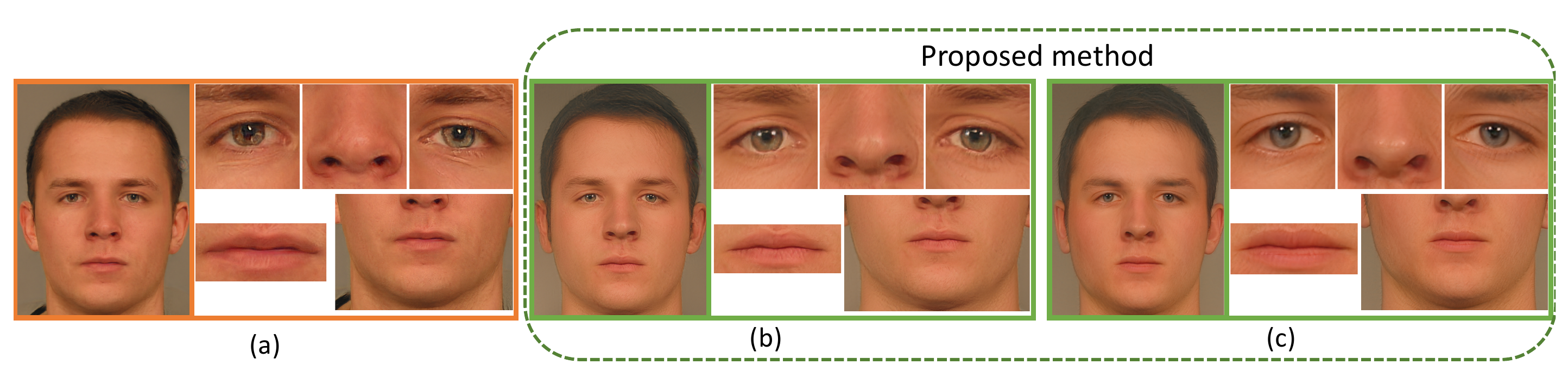}
	\caption{Details of segmented components in morphs generated by earlier method based on StyleGAN \cite{MorphStyleGAN2020} and proposed MIPGAN (a) StyleGAN \cite{MorphStyleGAN2020} (b) MIPGAN-I (c) MIPGAN-II.}
	\label{fig:component_details}
\end{figure*}
\subsection{Related Works on Face Morph Generation}
While morphing attacks have been studied in recent years, most of the attacks are conducted using the morphed images created using facial landmarks-based approaches needing high a degree of supervision to first determine the facial landmarks, thereupon align them and then finally blend them to generate morphs. The common set of procedures for warping/blending includes Free Form Deformation (FFD) \cite{lee1995ImgMetamorph} \cite{Warp1992feature}, Deformation by moving least squares  \cite{ImageDeformation-Schaefer2006}, deformation based on mass spring \cite{ImgMorph-MassSpring-choi2011},  Bayesian framework based morphing \cite{Interpolation-morping-Bichsel1996} and Delaunay triangulation based morphing \cite{visapp17_Morph} \cite{FRusingMOrphwu2011} \cite{IWBF2017_StirTrace} \cite{seibold2017ONLYVGGdetection} \cite{Scher2017}. Due to inadvertent artefacts caused by pixel/region-based morphing, the images need additional work in refining the signal to create highly realistic morph images. A set of post processing steps are usually included as illustrated in number of works \cite{seibold2017ONLYVGGdetection} \cite{FAceSwap-Bitouk2008} \cite{HairInterpolation-Weng2013}. Generally, some set of post processing techniques such as image smoothing, image sharpening, edge correction, histogram equalisation,  manual retouching, image enhancement to improve the brightness and contrast are used to eliminate the artefacts generated during the morphing process. In a parallel direction, morphed face images can also be generated using landmarks-based methods available in open-source resources like GIMP/GAP and OpenCV. Morphs generated using GIMP/GAP technique are more efficient with respect to a good quality of the resulting image (i.e., less noticeable artefacts) as pixels are aligned manually. Despite the minimal amount of effort needed for creating morphs using such approaches, a significant amount of effort needs to be dedicated to correcting artefacts. Additionally, commercial solutions like Face Fusion \cite{scherhag2019face} and FantaMorph \cite{FantaMorph} can also generate good quality morphed images with limited manual intervention. Although some steps can be excluded in creating the morphs, it is very critical to meet the face image quality standards laid out by the  International Civil Aviation Organization (ICAO) \cite{ICAO-9303-p1-2015}\cite{ICAO-9303-p9-2015} for electronic Machine Readable Travel Document (eMRTD) and deployment of biometric identification applications.
\subsection{GAN Based Face Morph Generation}
In an attempt to overcome the cumbersome efforts of manually creating (semi-automated) morphed images, a fully automated approach using a Generative Adversarial Network (GAN) was proposed by Damer et al.\cite{MorGAN}. Unlike the supervision required in the mark-up of landmarks and aligning the face images in a (partially) manual process, GAN-based techniques synthesise morphed images directly by merging two facial images in the latent space. In the work by Damer et al.\cite{MorGAN}, the proposed MorGAN architecture for morph generation basically employed a generator constituting encoders, decoders and a discriminator. The generator was trained to generate images with the dimension $64\times64$ pixels which is a key limiting factor of the attack, as most commercial FRS will reject images that do not meet the ICAO standard that requires a minimum Inter-Eye Distance (IED) of 90 pixels. The empirical evaluation of generated morph images using MorGAN in a vulnerability analysis against two commercial FRS indicated that those MorGAN morphs fail to meet both quality standards and the verification threshold of the FRS \cite{MorphStyleGAN2020}. Motivated to address the deficiency of the MorGAN architecture, in our recent work \cite{MorphStyleGAN2020} \footnote{The preliminary work results were published at IWBF-2020 in April, 2020.} we proposed an approach based on the StyleGAN architecture \cite{karras2019style} to increase the spatial dimension to $1024\times1024$ and thus to improve the face image quality. Unlike the previous approach of MorGAN \cite{MorGAN}, StyleGAN \cite{MorphStyleGAN2020} achieves better spatial resolution by embedding the images in the intermediate latent space. With the increased spatial dimension of resulting morphed images from our recently proposed architecture, we not only demonstrated that the images meet quality standards but also have a reasonable success rate when attacking commercial FRS \cite{MorphStyleGAN2020}. 

\subsection{Limitations of GAN Based Face Morph Generation and Our Contributions}
While our earlier work \cite{MorphStyleGAN2020} indicated that better GAN architectures could result in superior quality morphs and could attack an FRS in general, we also acknowledge the limited threats that exist for Commercial-Off-The-Shelf (COTS) FRS,  as merely a subset of morphed images was accepted. Only approximately $50\%$ of the generated morph images were verified successfully against probe images from a contributing subject. Thus the empirical evaluation in our earlier work has shown that the attack was yet not very effective \cite{MorphStyleGAN2020} 
for a COTS FRS\cite{cognitecfrssdk} and an open-source FRS based on ArcFace \cite{deng2019arcface}. We must state that up to now FRS are not very vulnerable to GAN-based morphing attacks unlike to landmarks-based morphing attacks. With a clear introspection into this aspect, we notice that the resulting morphed images from our earlier work \cite{MorphStyleGAN2020} does not retain a high degree of facial similarity to both contributing subjects. 
With lower similarity to contributing subjects in terms of facial structures, the FRS do not attribute a high comparison score, as anticipated. In other words, the missing enforcement of identity information of contributing subjects will lead to a high visual quality facial image but with lower face similarity to contributing face characteristics.

In an effort to make the attacks stronger such that both subjects can be verified with a good success rate, in this work, we extend our previous architecture to generate morphs by including the identity priors before the generation of morphed faces. We now refer to this approach as \textit{MIPGAN (Morphing through Identity Prior driven GAN)}. We propose two variants of our approach named as MIPGAN-I and MIPGAN-II based on the employed GAN being StyleGAN or StyleGAN2 respectively \cite{karras2019style, karras2020analyzing}. With the inclusion of a new loss function in our proposed architecture, we increase the attack success rate against commercial-off-the-shelf (COTS) FRS and deep learning based FRS. Figure~\ref{fig:iwbf2020-stylegan-morph-image} shows the example of morphed face images generated using proposed MIPGAN along with outputs of both the variants. To further achieve superior quality face morphs, we also customize the newly designed loss function to account for ghosting and blurring artefacts in an end-to-end manner with no human or manual intervention eliminating the need for a high degree of interaction. As noted in Figure~\ref{fig:component_details}, the results from MIPGAN-I and MIPGAN-II  is more coherent in retaining structural similarity as compared to our earlier architecture \cite{MorphStyleGAN2020}. With the updated architecture to generate high-quality morphs which preserve both identity information and structural correspondence, we evaluate the applicability in creating stronger attacks by creating a large-scale dataset of morphed images by employing the face images derived from the FRGC-V2 face database \cite{FRGC_DB}. The created dataset of 1270 bona fide images and 2500 morphed images is first evaluated to measure the attack success rate by verifying the morphed images against the contributing subjects using a commercial FRS from Cognitec \cite{cognitecfrssdk}. In addition to measuring the attack success rate for digital images, we also extend our work by printing and scanning (re-digitizing) the dataset. We check the consistency of the attack success rate, unlike our earlier work which was limited to an investigation on digital images alone \cite{MorphStyleGAN2020}. We also include the experiments on assessing the impact of compression (down to 15kb following ICAO guidelines) of printed and scanned face images that simulate the real-life e-passport application scenario. The key motivation to extend our work in this direction is, to mimic the passport application process that is operated in many European countries and Asian countries, which all accept printed-and-scanned facial images in the application process for an identity document (e.g. passports). 

With the extensive experimental results indicating a highly satisfactory attack success rate, we also evaluate a set of MAD algorithms to benchmark the detection capabilities. To this extent, we evaluate two state-of-the-art MAD approaches on digital morphed images, re-digitized and compressed morphed images after re-digitizing. Thus, we comprehensively cover the potential morphing attacks in the digital domain and the re-digitized domain. While we note the earlier works \cite{MorphStyleGAN2020} arguing that attacks in the digital domain can be detected by studying the cues such as residual noise in morphing \cite{Sushma_IPTA2019}, patterns of noise from morphed images, histogram features of textures or the deep features \cite{raghavendra2017transferable}, we also investigate the MAD capabilities for re-digitized images which do not exhibit the similar features (residual noise) as the print-scan process eliminates the digital cues and presents another set of variations. Specifically, given the nature of the dataset in which we have only a single suspected morphed image, for which we must determine either the morph or the bona fide class, we resort to Single Image based MAD (S-MAD) approaches using two recent but robust approaches using hybrid and ensemble features \cite{RagISBA2019, Sushma_IPTA2019, EnsembleFeatures_2020, NistFrvtMorph}.

\begin{figure*}[t!]
	\centering
	\includegraphics[width=1\linewidth]{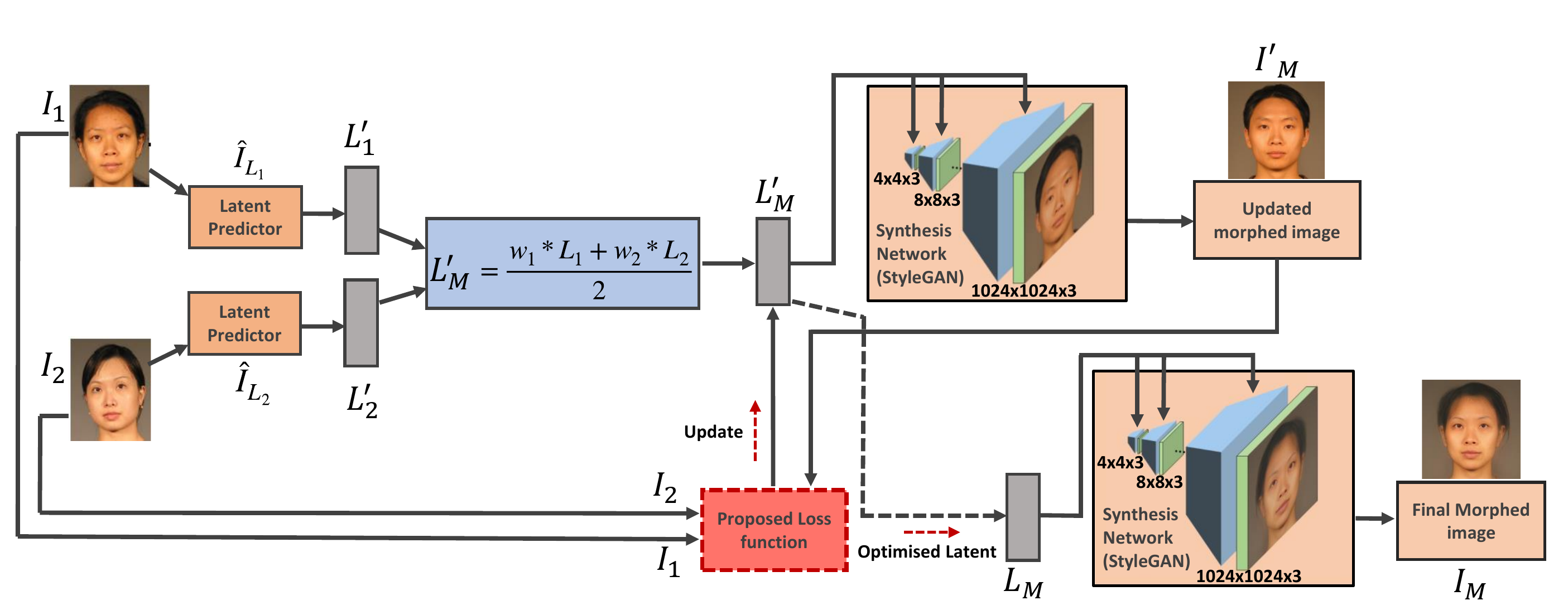}
	\caption{Block diagram of the proposed MIPGAN for generating high quality morphed face images}
	\label{fig:proposed_method}
\end{figure*}

We therefore present a summary of contributions of this work as listed below:
\begin{itemize}
    \item We present a novel approach of generating morphed face images through GAN architecture with enforced identity priors and a customized novel loss function to generate highly realistic images which we refer as \textit{MIPGAN (Morphing through Identity Prior driven GAN).} We present two variants of the proposed approach for generating attacks with a high success rate.
    \item The proposed approach (both variants) is benchmarked to measure the attack success rate by verifying COTS and deep learning based FRS through studying the vulnerability using a newly generated dataset from our proposed architecture which is referred as \textit{MIPGAN Face Morph Dataset}. 
    \item Human observer analysis for detecting morphs generated by the proposed and existing morphing attack methods is presented. 
    \item Analysis of the perceptual quality metrics to illustrate the visual quality of the generated morph images is presented.
    \item Extensive experiments on three different data types such as (a) digital morphed images (b) print-scan morphed image (c) print-scan morphed images with compression are presented to cover the full spectrum of passport application process under morphing attacks.
    \item The generated images are also benchmarked against the existing MAD approaches both in digital form and the re-digitized form to provide the insights on detection challenges of SOTA approaches. We also present a generalizability study on MAD schemes by training one kind of morph generation and testing on a different kind of morph generation approach to indicate directions to future works.

\end{itemize}

In the rest of the paper, Section \ref{sec:proposed-morph-generation} describes the new architecture along with the newly designed loss function to generate high-quality morphs. Section \ref{sec:experiments-results} provides the details on the quantitative experiments indicating the vulnerability of FRS and the detection challenge. With the set of remarks and future works in this direction, we draw the conclusion in Section~\ref{sec:conclusion}.

\section{Proposed Morphed Face Generation}
\label{sec:proposed-morph-generation}

Figure \ref{fig:proposed_method} presents the block diagram of the proposed morphed face image generation using MIPGAN. The proposed method is based on the end-to-end optimisation using a new loss function that can preserve the identity of the generated morphed face image through enforced identity priors.  The proposed MIPGAN framework is designed independently on two different GAN models based on StyleGAN \cite{karras2019style} and StyleGAN2 \cite{karras2020analyzing} model.  We refer to the proposed scheme with StyleGAN as MIPGAN-I and with StyleGAN2 as MIPGAN-II respectively. 
Given the face images from the accomplice ($I_{1}$) (contributing subject 1) and the malicious ($I_{2}$) (contributing subject 2)  data subjects, we predict the corresponding latent vectors $L_{1}^{'}$ and $L_{2}^{'}$ in the first step. 
In this work, we have employed the open-source pre-trained prediction models trained to predict the corresponding latent vector given an input image. Hence, $L’_1$ and $L’_2$ are predictions from the final output layer of the model, which is further reshaped.  Since  MIPGAN-I and MIPGAN-II are based on pre-trained StyleGAN \cite{karras2019style} and StyleGAN2 \cite{karras2019analyzing} model respectively, we used two different open-source pre-trained models for prediction. Both of the prediction models employ $ResNet50$ \cite{he2016deep} as backbone. The model for MIPGAN-I (StyleGAN) uses one convolution layer and two tree-connected layers \cite{richter2018treeconnect} to map the output of $ResNet50$ into the final latent vector with the size of $(18,512)$. In comparison, the model for MIPGAN-II (StyleGAN2) just uses one fully-connected layer to achieve the mapping. The predicted latent vectors thus provide the initialization for the morphed face generation that is obtained using a weighted linear average of $L_{1}^{'}$  and  $L_{2}^{'}$ as follows:
\begin{equation}
	L^{'}_{M} = \frac{w_{1}*L_{1}^{'} + w_{2}*L_{2}^{'}}{2},
	\label{eqn:Mor}
\end{equation}
where $w_{1}$ and $w_{2}$ indicate the weights, which we have chosen to be $w_{1} = w_{2} = 1$.  Equal weights are selected as shown in earlier work \cite{Venkatesh_2020_IJCB}  where the morphing images generated with equal weights pose higher vulnerability to COTS FRS. Finally, $L^{'}_{M}$ is passed through the synthesis network (independently from StyleGAN \cite{karras2019style} and StyleGAN2 \cite{karras2020analyzing} model) to generate the corresponding morphed image $I^{'}_{M}$ that has a resolution of $1024 \times 1024$ pixels. The generated morphed face image $I^{'}_{M}$ is then optimised using the proposed loss function to generate the high quality morphed face image. In the following section, we discuss the loss function to optimise the latent vector obtained using Equation \ref{eqn:Mor}. 
\subsection{Proposed Loss Function}
The proposed loss function is based on both perceptual fidelity, quality and identity factors that can facilitate high-quality face morph generation. The common issue with the GAN-based morph generation is the presence of ghost artefacts and blurring issues. We employ the perceptual loss with multiple layers to eliminate such effects as given by Eqn~\ref{eqn:loss-perceptual}.
\begin{equation}
\begin{aligned}
   Loss_{Perceptual} &= \frac{1}{2}\sum_{i} \frac{1}{N_i}||F_i(I_1)-F_i(I^{'}_{M})||^2_2 \\
                    &\quad + \frac{1}{2}\sum_{i} \frac{1}{N_i}||F_i(I_2)-F_i(I^{'}_{M})||^2_2,
\end{aligned}
\label{eqn:loss-perceptual}
\end{equation}
where $N_i$ denotes the number of features in layer $i$ and $F_i$ denotes features in layer $i$ of the perceptual network (VGG-16 in our case). For the combination of perceptual layers, we choose $conv1_1$, $conv1_2$, $conv2_2$, $conv3_3$ inspired by \cite{abdal2019image2stylegan++}. Compared with the original combination of layers $conv1_2$, $conv2_2$, $conv3_3$, $conv4_3$ \cite{johnson2016perceptual}, our design measures  low-level features instead of high-level features like style of an image and is closer to our goal of morphing faces with high quality. 

The main goal of this paper is to generate the morphed face images that can significantly attack FRS. In order to achieve this, we have introduced the identity loss function based on the feedback from FRS. We employ Arcface \cite{deng2019arcface} - a deep learning based FRS because of its robust and accurate performance to obtain feedback on generated morphed face images. Specifically, we employ a pre-trained embedding extractor with $ResNet50$ as the backbone to extract the unit embedding vectors and define the identity loss by their cosine distance to improve the morph generation process as given by Eqn~\ref{eqn:identity-loss}.
\begin{equation}
    Loss_{Identity}=\frac{(1-\frac{\Vec{v}_1 \cdot \vec{v}_M}{\Vert\vec{v}_1\Vert \Vert\vec{v}_M\Vert})+(1-\frac{\vec{v}_2 \cdot \vec{v}_M}{\Vert\vec{v}_2\Vert\Vert\vec{v}_M\Vert})}{2},
    \label{eqn:identity-loss}
\end{equation}
where $\vec{v}_1$, $\vec{v}_2$, $\vec{v}_M$ respectively denotes the embedding vectors which are extracted from image $I_1, I_2, I^{'}_{M}$ respectively.  

To further prove the loss function is differential for the morphed embedding vector $\vec{v}_M$, we define $x_d, y_d, z_d$ to be the value of vector $\vec{v}_1$, $\vec{v}_2$, $\vec{v}_M$ in dimension $d$ respectively and $d' \neq d$ to be other dimensions except $d$. The expanded identity loss function and its partial derivative are: 
\begin{equation}
    Loss_{Identity}=\frac{(1-\frac{\sum_d x_d z_d}{\Vert\vec{v}_1\Vert\Vert\vec{v}_M\Vert})+(1-\frac{\sum_d y_d z_d}{\Vert\vec{v}_2\Vert\Vert\vec{v}_M\Vert})}{2},
\end{equation}

\begin{equation}
    \begin{aligned}
    \frac{\partial Loss_{Identity}}{\partial z_d} &= 1-\frac{x_d}{2\Vert\vec{v}_1\Vert}\frac{\partial }{\partial z_d}(\frac{z_d}{\sqrt{z_d^2+\sum_{d'\neq d}z_{d'}^2}}) \\
                                                  &\quad - \frac{y_d}{2\Vert\vec{v}_2\Vert}\frac{\partial }{\partial z_d}(\frac{z_d}{\sqrt{z_d^2+\sum_{d'\neq d}z_{d'}^2}}),
    \end{aligned}
\end{equation}

\begin{align*}
    \frac{\partial}{\partial z_d}(\frac{z_d}{\sqrt{z_d^2+\sum_{d'\neq d}z_{d'}^2}}) &= \frac{1}{\sqrt{z_d^2+\sum_{d'\neq d}z_{d'}^2}}\\
                                                                                    &\quad +\frac{2z_d^2}{-2(z_d^2+\sum_{d'\neq d}z_{d'}^2)^{\frac{3}{2}}}\\
    &=\frac{\sum_{d'\neq d}z_{d'}^2}{(z_d^2+\sum_{d'\neq d}z_{d'}^2)^{\frac{3}{2}}},
\end{align*}
\begin{equation}
    \frac{\partial Loss_{Identity}}{\partial z_d}=1-\frac{(\frac{x_d}{2\Vert\vec{v}_1\Vert}+\frac{y_d}{2\Vert\vec{v}_2\Vert})\sum_{d'\neq d}z_{d'}^2}{(z_d^2+\sum_{d'\neq d}z_{d'}^2)^{\frac{3}{2}}}.
\end{equation}
For any value $z_d=z'_{d}$, it is obvious that:
\begin{align*}
    &\lim_{\Delta z_d \to 0}\frac{\partial Loss_{Identity}(z'_d+\Delta z_d)}{\partial z_d}\\
    &=\lim_{\Delta z_d \to 0}(1-\frac{(\frac{x_d}{2\Vert\vec{v}_1\Vert}+\frac{y_d}{2\Vert\vec{v}_2\Vert})\sum_{d'\neq d}z_{d'}^2}{((z'_{d}+\Delta z_d)^2+\sum_{d'\neq d}z_{d'}^2)^{\frac{3}{2}}})\\
    &=1-\frac{(\frac{x_d}{2\Vert\vec{v}_1\Vert}+\frac{y_d}{2\Vert\vec{v}_2\Vert})\sum_{d'\neq d}z_{d'}^2}{(z^{'2}_d+\sum_{d'\neq d}z_{d'}^2)^{\frac{3}{2}}}\\
    &=\frac{\partial Loss_{Identity}(z'_d)}{\partial z_d}.
\end{align*}
Hence, for any dimension of $d$, the partial derivative of the identity loss function is continuous. 

\begin{figure*}[!b]
	\centering
	\includegraphics[width=1\linewidth]{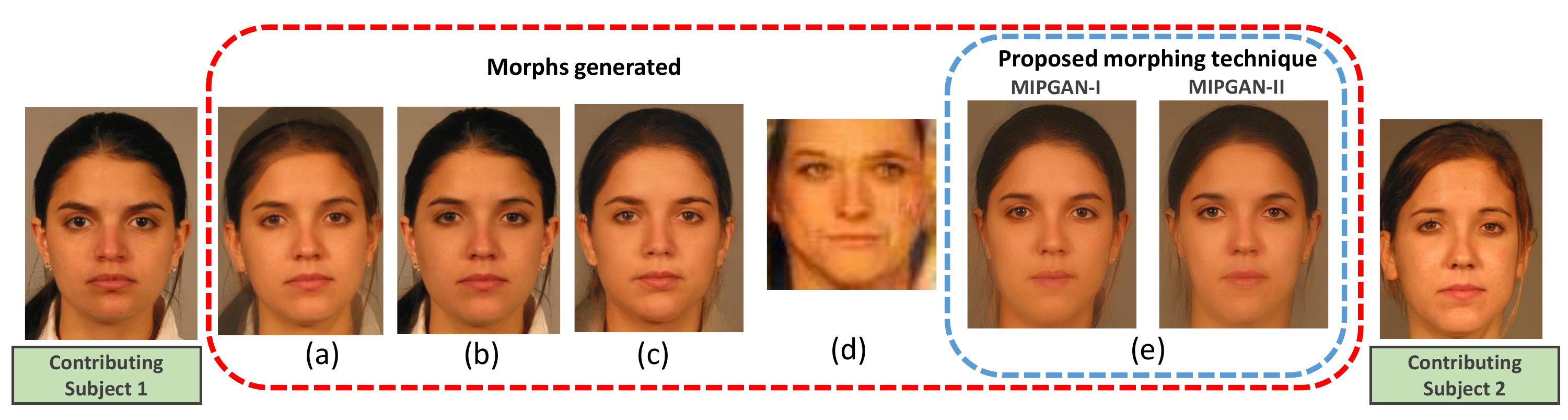}
	\caption{Qualitative results of proposed MIPGAN together with existing GAN based face morph generation methods (a) Landmark-I \cite{raghavendra2017face}  (b) Landmark-II \cite{UBO_Morphing_Tool} (c)  StyleGAN\cite{MorphStyleGAN2020} (d) MorGAN \cite{MorGAN} (e) Proposed method}
	\label{fig:MIPGANQualitativeRes}
\end{figure*}

It is interesting to note that the identity loss based on the Arcface feature extractor model is trained to maximize the face class separability and thus is more sensitive to face attributes. Hence, only optimising the identity loss cannot achieve the same reconstruction performance as the perceptual loss but applying it on the face region can effectively control the generated attributes to be recognized as both subjects. 

To solve the imbalance between different subjects, we introduce an identity difference loss as given by Eqn~\ref{eqn:identity-diff-loss}.
\begin{equation}
    Loss_{ID-Diff}=|(1-\frac{\vec{v}_1\cdot\vec{v}_M}{\Vert\vec{v}_1\Vert\Vert\vec{v}_M\Vert}) -(1-\frac{\vec{v}_2\cdot\vec{v}_M}{\Vert\vec{v}_2\Vert\Vert\vec{v}_M\Vert})|.
    \label{eqn:identity-diff-loss}
\end{equation}
With the idea of the Lagrange multiplier, it adds a constraint to the optimisation process to force the cosine distance between morph embedding and each of the two reference embeddings to be the same. Since $Loss_{ID-Diff}$ is usually small with a value less than $1$, we apply $L1$ loss on the difference of two cosine distance terms to avoid the vanishing gradient problem. 

Finally, in order to improve the structural visibility of the generated morphed face image, we also apply the Multi-Scale Structural Similarity (MS-SSIM) loss $L_{MS-SSIM}$ to measure the similarity in structure \cite{wang2003multiscale}. Given two discrete non-negative signals (images in our case) $x$ and $y$, luminance, contrast and structure
comparison measures were given  by $l, c, s$ as computed using Eqn~\ref{eqn:luminance-contrast-structure}.
\begin{equation}
\begin{aligned}
    l(x,y) &=\frac{(2\mu_x2\mu_y+(K_1L)^2)}{\mu_x^2+\mu_y^2+(K_1L)^2},\\
    c(x,y) &=\frac{(2\sigma_x2\sigma_y+(K_2L)^2)}{\sigma_x^2+\sigma_y^2+(K_2L)^2},\\
    s(x,y) &=\frac{(\sigma_{xy}+\frac{(K_2L)^2}{2})}{\sigma_x\sigma_y+\frac{(K_2L)^2}{2}},
    \end{aligned}
    \label{eqn:luminance-contrast-structure}
\end{equation} 
where $\mu_x, \sigma_x$ and $\sigma_{xy}$ denotes the mean of $x$, the variance of $x$ and the covariance of $x$ and $y$ respectively. $L$ is the dynamic range of the signal and $K_1 \ll 1, K_2 \ll 1$ are two constant scalars. The MSSSIM loss $L_{MS-SSIM}$ is further defined by Eqn~\ref{eqn:loss-ms-ssim}.
\begin{equation}
\begin{aligned}
    MSSSIM(x,y) =& [l_J(x,y)]^{\alpha_J} \cdot \prod_{j=1}^J [c_j(x,y)]^{\beta_j}[s_j(x,y)]^{\gamma_j},\\
    L_{MS-SSIM} =& \frac{1}{2}(1-MSSSIM(I_1,I'_M))\\
    &+\frac{1}{2}(1-MSSSIM(I_2,I'_M)),
    \end{aligned}
    \label{eqn:loss-ms-ssim}
\end{equation}
where $j=1,2,\ldots,J$ represents the $j^{th}$ scale and $\alpha_j,\beta_j$ and $\gamma_j$ are the factors of relative importance. As suggested in  \cite{wang2003multiscale}, we also set $\alpha_j = \beta_j = \gamma_j$, $\sum_{j=1}^J\gamma_j=1$ and use the resulting parameters $\beta_1=\gamma_1=0.0448, \beta_2=\gamma_2=0.2856, \beta_3=\gamma_3=0.3001, \beta_4=\gamma_4=0.2363, \alpha_5=\beta_5=\gamma_5=0.1333$.

Thus, the proposed loss function can be formulated as:
\begin{equation}
\begin{aligned}
    Loss &=\lambda_1 Loss_{Perceptual}+\lambda_2 Loss_{Identity}\\
        &\quad+\lambda_3 Loss_{MS-SSIM} + \lambda_4 Loss_{ID-Diff},
    \end{aligned}
    \label{eqn:Finalloss}
\end{equation} 
where $\lambda_1$, $\lambda_2$, $\lambda_3$ and $\lambda_4$ are the hyper-parameters that are set to achieve both stable and generalised convergence. In this work, we empirically set  $\lambda_1 = 0.0002$, $\lambda_2 = 10$, $\lambda_3 = 1$ and $\lambda_4 = 1$. 

\subsection{Training and Optimisation}
The training and optimisation of the proposed method are carried out on Tensorflow version 1.13 and version 1.14 for StyleGAN and StyleGAN2, respectively. The optimisation is carried out using NVIDIA GTX 1070 8 GB GPU with CUDA version 10.0 and CUDNN version 7.5 and NVIDIA Tesla P100 PCIE 16 GB GPU. The Adam optimiser with hyper-parameters $\beta_1=0.9$, $\beta_2=0.999$ and $\epsilon=1\times10^{-8}$ as recommended in the original paper \cite{kingma2014adam} is employed on this work. The list of morphing pairs is generated in advance with careful considerations to gender. During each optimisation process of 150 iterations, the learning rate is initially set to $\eta = 0.03$ with an exponential decay per 6 iterations of $\eta_{new}=\eta*0.95$.

Figure \ref{fig:MIPGANQualitativeRes} illustrates the qualitative results of the proposed MIPGAN framework based on StyleGAN and StyleGAN2.  Further, the qualitative results of the existing methods based on StyleGAN \cite{MorphStyleGAN2020} and MorGAN \cite{MorGAN} are provided alongside for the convenience of the reader in the same figure.  It is interesting to note that the proposed MIPGAN generated face morph images indicate both perceptual and geometric features correspondence to both contributing subjects (for instance, malicious actor and accomplice). 
\section{Experiments and results}
\label{sec:experiments-results}
This section presents and discusses the experimental protocols, datasets, and quantitative results of the proposed face morphing technique. The images generated from the proposed MIPGAN-I and MIPGAN-II architectures are compared with the state-of-the-art techniques based on both facial landmarks \cite{raghavendra2017face} and StyleGAN based morph generation \cite{MorphStyleGAN2020}. The effectiveness of the face morphing generation is quantitatively evaluated by benchmarking the vulnerability of the COTS FRS and deep learning based FRS for generated morphed face images. Further, we also evaluate the morphing attack detection potential by evaluating the generated morphed face images using the most recent and robust MAD techniques.

\begin{figure*}[htp]
	\centering
	\includegraphics[width=1\linewidth]{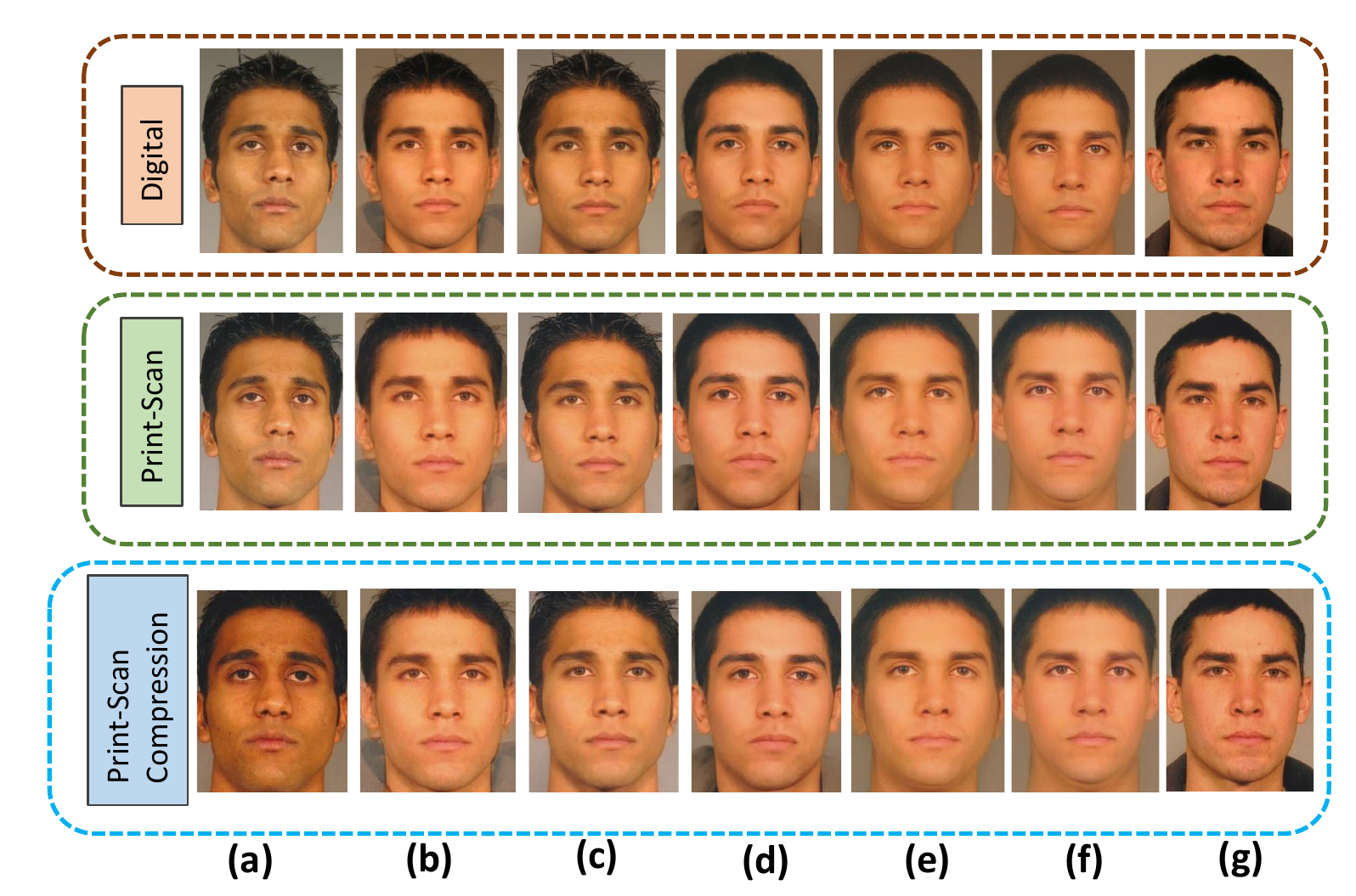}
	\caption{Illustration of morphing in digital, print-scan and print-scan compression data (a) Contributing subject 1 (b) Landmark-I  \cite{raghavendra2017face} (c) Landmark-II \cite{UBO_Morphing_Tool} (d)  StyleGAN \cite{MorphStyleGAN2020} (e) MIPGAN-I (f) MIPGAN-II (g)  Contributing subject 2}
	\label{fig:illusPrintScan}
\end{figure*}

\subsection{MIPGAN Face Morph Dataset}
We employ the face images from FRGC-V2 face database \cite{FRGC_DB} to generate the \textit{MIPGAN Face Morph Dataset} consisting of morphed face images using both state-of-the-art and the proposed MIPGAN technique. We have selected 140 unique data subjects from the FRGC dataset by considering the high-quality face images captured in constrained conditions that resemble the passport image quality. Among 140 data subjects, 47 data subjects are female and 93 data subjects are male. Each data subject has a variable size of 7-21 additional captured samples, resulting for the whole dataset to have 1270 samples corresponding to 140 data subjects. We employ three different face morph generation techniques based on facial landmarks constrained by Delaunay triangulation with blending \cite{raghavendra2017face} we term this as Landmarks-I, landmarks-based techniques with automatic post processing and color equalisation \cite{UBO_Morphing_Tool}, we term this as Landmarks-II and StyleGAN  \cite{MorphStyleGAN2020}. We do not consider MorGAN \cite{MorGAN} \cite{realistic_dreams_MorGAN} based face morph generation as it was earlier demonstrated that MorGAN does not generate ICAO compliant images and thus makes COTS FRS not vulnerable \cite{MorphStyleGAN2020}.  All the samples are pre-processed to meet the ICAO standards \cite{ICAO-9303-p9-2015} and morphing is carried out by following the guidelines outlined earlier \cite{raghavendra2017face} \cite{scherhag2017biometric}, i.e, careful selection of subjects based on gender and similarity score using a FRS, in order to have realistic attacks. 

To effectively evaluate the proposed method's quantitative performance and the existing techniques, we create three different types of attacks from morphed images, such as \textbf{Digital morphed images}: Morphed face images that are obtained from the morph generation process in the digital domain. \textbf{Print-scanned morphed images}: The digital morphed and bona fide images are printed and then scanned (or re-digitized) to simulate the passport application process. We have employed a DNP-DS820 \cite{DNP_Printer} dye-sublimation photo printer to generate the prints of the digital morphed and bona fide face images in this work. The use of a dye-sublimation photo printer guarantees high-quality photo printing (generally used for a passport application) and makes sure that printed photos are free from dotted patterns (or individual droplets of ink) that are resulting from the printing process of conventional printers. Each of these printed photos is then scanned (or re-digitized) using the Canon office scanner to have 300 dpi as suggested in ICAO standards \cite{ICAO-9303-p9-2015}. \textbf{Print-scanned compressed morphed images}: The printed and scanned images (both morphed and bona fide) are compressed to have a size of 15kb that makes it suitable to store in the e-passport. This process reflects the real-life scenario of face image storage in passport systems. Thus, the overall dataset has $2500$ $\times$ $3$ (types of morph data) $\times$ $4$ types of morph generation technique = $30,000$ morph samples and $1270$ $\times$ $3$ (types of morph data) $\times$ $4$ types of morph generation technique = $15,240$ bona fide samples. Figure \ref{fig:illusPrintScan} illustrates the three data types of attacks that are used to evaluate the effectiveness of the proposed method and the existing methods of face morph generation. It is evident that the visual quality of the images vary largely for different attack types (for instance, the digital data attack indicates the best quality and print-scan with compression indicates the lowest quality).   
\begin{table*}[htbp]
  \centering

    \caption{Quantitative evaluation of vulnerability of COTS Cognitec-FRS \cite{cognitecfrssdk} from various morph generation approaches. Note that, since FNMR = $0$ $@$ FMR = $0.1\%$ for Cognitec-FRS \cite{cognitecfrssdk} following Eq. \ref{Eqa:RMMR_MMPMR} and \ref{Eqa:RMMR_FMMPMR}, the value of RMMR is equal to MMPMR/FMMPMR. Therefore, we have not entered RMMR separately in the Table above.}
    \resizebox{1\linewidth}{!}{
    \begin{tabular}{|c|c|c|c|c|c|c|}
    \hline
    \cline{2-7}    \multirow{3}[6]{*}{\textbf{Morph generation type\newline{}\newline{}}} & MMPMR/RMMR(\%) & FMMPMR/RMMR(\%) & MMPMR/RMMR(\%) & FMMPMR/RMMR(\%) & MMPMR/RMMR(\%) & FMMPMR/RMMR(\%) \bigstrut\\
\cline{2-7}    \multicolumn{1}{|c|}{} & \multicolumn{2}{c|}{\textbf{Digital}} & \multicolumn{2}{c|}{\textbf{Print-Scan}} & \multicolumn{2}{c|}{\textbf{Print-Scan with compression}} \bigstrut\\
\cline{2-7}    \multicolumn{1}{|c|}{} & \multicolumn{2}{c|}{Male} & \multicolumn{2}{c|}{Male} & \multicolumn{2}{c|}{Male} \bigstrut\\
    \hline
    Landmark-I \cite{raghavendra2017face} & \multicolumn{1}{c|}{100} & 98.77 & 97.23 & 97.34 & 97.38 & 96.95 \bigstrut\\
    \hline
    Landmark-II \cite{UBO_Morphing_Tool} & 87.29 & 76.86 & 90.32 & 78.23 & 88.78 & 77.14 \bigstrut\\
    \hline
    StyleGAN \cite{MorphStyleGAN2020} & 63.51 & 41.27 & 60.59 & 39.51 & 57.12 & 35.05 \bigstrut\\
    \hline
    MIPGAN-I & 93.35 & 83.08 & 91.72 & 80.55 & 91.07 & 77.89 \bigstrut\\
    \hline
    MIPGAN-II & 92.22 & 80.45 & 90.74 & 77.67 & 89.16 & 73.47 \bigstrut\\
    \hline
    \multicolumn{1}{|r|}{} & \multicolumn{2}{c|}{Female} & \multicolumn{2}{c|}{Female} & \multicolumn{2}{c|}{Female} \bigstrut\\
    \hline
    Landmark-I \cite{raghavendra2017face} & \multicolumn{1}{c|}{100} & 99.26 & 99.37 & 99.02 & 99.78 & 99.24 \bigstrut\\
    \hline
    Landmark-II \cite{UBO_Morphing_Tool} & 94.28 & 88.67 & 98.22 & 91.48 & 98.16 & 90.97 \bigstrut\\
    \hline
    StyleGAN \cite{MorphStyleGAN2020} & 68.75 & 42.62 & 66.45 & 42.01 & 66.45 & 40.49 \bigstrut\\
    \hline
    MIPGAN-I & 98.57 & 93.11 & 98.16 & 91.22 & 96.12 & 90.52 \bigstrut\\
    \hline
    MIPGAN-II & 95.91 & 87.66 & 95.30 & 86.26 & 94.69 & 84.47 \bigstrut\\
    \hline
    \multicolumn{1}{|r|}{} & \multicolumn{2}{c|}{Combined} & \multicolumn{2}{c|}{Combined} & \multicolumn{2}{c|}{Combined}\bigstrut\\
    \hline
    Landmark-I \cite{raghavendra2017face} & \multicolumn{1}{c|}{\textbf{100}} & \textbf{98.84} & \textbf{97.64} & \textbf{97.60} & \textbf{97.84} & \textbf{97.30} \bigstrut\\
    \hline
    Landmark-II \cite{UBO_Morphing_Tool} & 88.65 & 78.72 & 91.85 & 81.56 & 90.61 & 79.33 \bigstrut\\
    \hline
    StyleGAN \cite{MorphStyleGAN2020} & 64.68 & 41.49 & 61.72 & 39.90 & 58.92 & 35.89 \bigstrut\\
    \hline
    MIPGAN-I & 94.36 & 84.65 & 92.97 & 82.23 & 92.29 & 79.88 \bigstrut\\
    \hline
    MIPGAN-II & 92.93 & 81.59 & 80.56 & 79.02 & 90.24 & 75.20 \bigstrut\\
    \hline
    \end{tabular}%
    }
  \label{tab:vulnerability-report}%
  
\end{table*}%

\begin{table*}[htbp]
  \centering
    \caption{Quantitative evaluation of vulnerability of VGGFace2 \cite{cao2018vggface2} FRS from various morph generation approaches. Note that, since FNMR = $0$ $@$ FMR = $0.1\%$ for VGGFace2 \cite{cao2018vggface2} following Eq. \ref{Eqa:RMMR_MMPMR} and \ref{Eqa:RMMR_FMMPMR}, the value of RMMR is equal to MMPMR/FMMPMR. Therefore, we have not entered RMMR separately in the Table above.}
    \resizebox{1\linewidth}{!}{
    \begin{tabular}{|c|c|c|c|c|c|c|}
    \hline
    \cline{2-7}    \multirow{3}[6]{*}{\textbf{Morph generation type\newline{}\newline{}}} & MMPMR/RMMR(\%) & FMMPMR/RMMR(\%) & MMPMR/RMMR(\%) & FMMPMR/RMMR(\%) & MMPMR/RMMR(\%) & FMMPMR/RMMR(\%) \bigstrut\\
\cline{2-7}    \multicolumn{1}{|c|}{} & \multicolumn{2}{c|}{\textbf{Digital}} & \multicolumn{2}{c|}{\textbf{Print-Scan}} & \multicolumn{2}{c|}{\textbf{Print-Scan with compression}} \bigstrut\\
\cline{2-7}    \multicolumn{1}{|c|}{} & \multicolumn{2}{c|}{Male} & \multicolumn{2}{c|}{Male} & \multicolumn{2}{c|}{Male} \bigstrut\\
    \hline
    Landmark-I \cite{raghavendra2017face} & \multicolumn{1}{c|}{85.59} & 70.80 & 83.91 & 68.20 & 83.86 & 67.73 \bigstrut\\
    \hline
    Landmark-II \cite{UBO_Morphing_Tool} & 63.27 & 46.55 & 63.12 & 46.37 & 63.72 & 46.80 \bigstrut\\
    \hline
    StyleGAN \cite{MorphStyleGAN2020} & 61.19 & 41.01 & 61.68 & 41.43 & 61.68 & 41.04 \bigstrut\\
    \hline
    MIPGAN-I & 76.96 & 59.24 & 76.96 & 57.16 & 76.07 & 57.31 \bigstrut\\
    \hline
    MIPGAN-II & 75.73 & 56.97 & 72.87 & 54.57 & 72.87 & 54.43 \bigstrut\\
    \hline
    \multicolumn{1}{|r|}{} & \multicolumn{2}{c|}{Female} & \multicolumn{2}{c|}{Female} & \multicolumn{2}{c|}{Female} \bigstrut\\
    \hline
    Landmark-I \cite{raghavendra2017face} & \multicolumn{1}{c|}{96.03} & 83.55 & 93.95 & 82.02 & 93.32 & 81.39 \bigstrut\\
    \hline
    Landmark-II \cite{UBO_Morphing_Tool} & 87.76 & 71.85 & 89.39 & 73.82 & 89.80 & 74.27 \bigstrut\\
    \hline
    StyleGAN \cite{MorphStyleGAN2020} & 80.42 & 59.19 & 79.79 & 59.10 & 78.54 & 58.83 \bigstrut\\
    \hline
    MIPGAN-I & 90.41 & 76.68 & 89.39 & 75.95 & 89.18 & 75.85 \bigstrut\\
    \hline
    MIPGAN-II & 88.98 & 75.42 & 87.96 & 74.54 & 88.37 & 74.90 \bigstrut\\
    \hline
    \multicolumn{1}{|r|}{} & \multicolumn{2}{c|}{Combined} & \multicolumn{2}{c|}{Combined} & \multicolumn{2}{c|}{Combined}\bigstrut\\
    \hline
    Landmark-I \cite{raghavendra2017face} & \multicolumn{1}{c|}{\textbf{87.64}} & \textbf{72.82} & \textbf{85.87} & \textbf{70.39} & \textbf{85.71} & \textbf{69.90} \bigstrut\\
    \hline
    Landmark-II \cite{UBO_Morphing_Tool} & 68.07 & 50.64 & 68.27 & 50.80 & 68.86 & 51.28 \bigstrut\\
    \hline
    StyleGAN \cite{MorphStyleGAN2020} & 64.92 & 43.91 & 65.20 & 44.25 & 64.96 & 43.88 \bigstrut\\
    \hline
    MIPGAN-I & 79.61 & 62.06 & 79.41 & 60.19 & 78.66 & 60.30 \bigstrut\\
    \hline
    MIPGAN-II & 78.34 & 59.95 & 75.84 & 57.80 & 75.92 & 57.73 \bigstrut\\
    \hline
    \end{tabular}%
    }
  \label{tab:vulnerability-report-VGGFace2}%
\end{table*}%

\begin{table*}[htbp]
  \centering
    \caption{Quantitative evaluation of vulnerability of Arcface \cite{deng2019arcface} FRS from various morph generation approaches. Note that, since FNMR = $0$ $@$ FMR = $0.1\%$ for Arcface \cite{deng2019arcface} following Eq. \ref{Eqa:RMMR_MMPMR} and \ref{Eqa:RMMR_FMMPMR}, the value of RMMR is equal to MMPMR/FMMPMR. Therefore, we have not entered RMMR separately in the Table above.}
    \resizebox{1\linewidth}{!}{
    \begin{tabular}{|c|c|c|c|c|c|c|}
    \hline
    \cline{2-7}    \multirow{3}[6]{*}{\textbf{Morph generation type\newline{}\newline{}}} & MMPMR/RMMR(\%) & FMMPMR/RMMR(\%) & MMPMR/RMMR(\%) & FMMPMR/RMMR(\%) & MMPMR/RMMR(\%) & FMMPMR/RMMR(\%) \bigstrut\\
\cline{2-7}    \multicolumn{1}{|c|}{} & \multicolumn{2}{c|}{\textbf{Digital}} & \multicolumn{2}{c|}{\textbf{Print-Scan}} & \multicolumn{2}{c|}{\textbf{Print-Scan with compression}} \bigstrut\\
\cline{2-7}    \multicolumn{1}{|c|}{} & \multicolumn{2}{c|}{Male} & \multicolumn{2}{c|}{Male} & \multicolumn{2}{c|}{Male} \bigstrut\\
    \hline
    Landmark-I \cite{raghavendra2017face} & \multicolumn{1}{c|}{99.60} & 98.19 & 97.38 & 96.88 & 97.33 & 96.70 \bigstrut\\
    \hline
    Landmark-II \cite{UBO_Morphing_Tool} & 91.09 & 84.62 & 93.45 & 86.42 & 93.60 & 86.02 \bigstrut\\
    \hline
    StyleGAN\cite{MorphStyleGAN2020} & 70.99 & 55.76 & 73.86 & 58.67 & 73.32 & 58.26 \bigstrut\\
    \hline
    MIPGAN-I & 93.70 & 85.17 & 92.76 & 84.39 & 93.01 & 84.41 \bigstrut\\
    \hline
    MIPGAN-II & 93.65 & 86.45 & 93.55 & 85.30 & 93.25 & 85.06 \bigstrut\\
    \hline
    \multicolumn{1}{|r|}{} & \multicolumn{2}{c|}{Female} & \multicolumn{2}{c|}{Female} & \multicolumn{2}{c|}{Female} \bigstrut\\
    \hline
   Landmark-I \cite{raghavendra2017face} & \multicolumn{1}{c|}{99.79} & 97.01 & 99.79 & 96.91 & 99.79 & 97.01 \bigstrut\\
    \hline
   Landmark-II \cite{UBO_Morphing_Tool} & 94.49 & 86.71 & 97.76 & 89.76 & 98.16 & 89.17 \bigstrut\\
    \hline
    StyleGAN \cite{MorphStyleGAN2020} & 80.21 & 63.22 & 82.71 & 65.70 & 82.71 & 66.05 \bigstrut\\
    \hline
    MIPGAN-I & 97.35 & 89.53 & 97.96 & 91.02 & 97.76 & 91.02 \bigstrut\\
    \hline
    MIPGAN-II & 96.33 & 89.47 & 95.92 & 89.33 & 96.12 & 89.42 \bigstrut\\
    \hline
    \multicolumn{1}{|r|}{} & \multicolumn{2}{c|}{Combined} & \multicolumn{2}{c|}{Combined} & \multicolumn{2}{c|}{Combined}\bigstrut\\
    \hline
   Landmark-I \cite{raghavendra2017face} & \multicolumn{1}{c|}{\textbf{99.68}} & \textbf{98.00} & \textbf{97.88} & \textbf{96.89} & \textbf{97.84} & \textbf{96.75} \bigstrut\\
    \hline
    Landmark-II \cite{UBO_Morphing_Tool} & 91.79 & 84.96 & 94.33 & 86.96 & 94.53 & 86.54 \bigstrut\\
    \hline
    StyleGAN \cite{MorphStyleGAN2020} & 72.80 & 56.95 & 75.60 & 59.79 & 75.16 & 59.51 \bigstrut\\
    \hline
    MIPGAN-I & 94.45 & 85.94 & 93.81 & 85.46 & 93.97 & 85.48 \bigstrut\\
    \hline
    MIPGAN-II & 94.21 & 86.94 & 94.05 & 85.95 & 93.85 & 85.77 \bigstrut\\
    \hline
    \end{tabular}%
    }
  \label{tab:vulnerability-report-Arcface}%
\end{table*}%

\begin{table*}[htbp]
  \centering
    \caption{Quantitative evaluation of vulnerability of COTS Neurotec \cite{Neurotech} FRS from various morph generation approaches. Note that, since FNMR = $0$ $@$ FMR = $0.1\%$ for COTS Neurotec \cite{Neurotech} following Eq. \ref{Eqa:RMMR_MMPMR} and \ref{Eqa:RMMR_FMMPMR}, the value of RMMR is equal to MMPMR/FMMPMR. Therefore, we have not entered RMMR separately in the Table above.}
    \resizebox{1\linewidth}{!}{
    \begin{tabular}{|c|c|c|c|c|c|c|}
    \hline
    \cline{2-7}    \multirow{3}[6]{*}{\textbf{Morph generation type\newline{}\newline{}}} & MMPMR/RMMR(\%) & FMMPMR/RMMR(\%) & MMPMR/RMMR(\%) & FMMPMR/RMMR(\%) & MMPMR/RMMR(\%) & FMMPMR/RMMR(\%) \bigstrut\\
\cline{2-7}    \multicolumn{1}{|c|}{} & \multicolumn{2}{c|}{\textbf{Digital}} & \multicolumn{2}{c|}{\textbf{Print-Scan}} & \multicolumn{2}{c|}{\textbf{Print-Scan with compression}} \bigstrut\\
\cline{2-7}    \multicolumn{1}{|c|}{} & \multicolumn{2}{c|}{Male} & \multicolumn{2}{c|}{Male} & \multicolumn{2}{c|}{Male} \bigstrut\\
    \hline
   Landmark-I \cite{raghavendra2017face} & \multicolumn{1}{c|}{99.40} & 94.70  & 95.45 & 83.71 & 93.23 & 77.16 \bigstrut\\
    \hline
    Landmark-II \cite{UBO_Morphing_Tool} & 88.99 & 68.51 & 88.92 & 63.31 & 80.62 & 53.63 \bigstrut\\
    \hline
    StyleGAN \cite{MorphStyleGAN2020} & 52.26 & 26.47 & 31.88 & 12.98 & 31.60  & 12.15 \bigstrut\\
    \hline
    MIPGAN-I & 58.18 & 32.56 & 32.59 & 25.33 & 57.6  & 53.52 \bigstrut\\
    \hline
    MIPGAN-II & 53.16 & 29.65 & 47.41 & 20.71 & 50.73 & 23.72 \bigstrut\\
    \hline
    \multicolumn{1}{|r|}{} & \multicolumn{2}{c|}{Female} & \multicolumn{2}{c|}{Female} & \multicolumn{2}{c|}{Female} \bigstrut\\
    \hline
    Landmark-I \cite{raghavendra2017face} & \multicolumn{1}{c|}{100} & 99.25 & 100   & 98.11 & 98.74 & 91.18 \bigstrut\\
    \hline
    Landmark-II \cite{UBO_Morphing_Tool} & 94.69 & 85.96 & 97.49 & 84.92 & 95.40  & 78.89 \bigstrut\\
    \hline
    StyleGAN \cite{MorphStyleGAN2020} & 70.60  & 50.13 & 55.20  & 25.72 & 52.39 & 26.19 \bigstrut\\
    \hline
    MIPGAN-I & 80.98 & 56.29 & 73.06 & 46.87 & 77.89 & 30.50 \bigstrut\\
    \hline
    MIPGAN-II & 74.79 & 49.45 & 69.59 & 42.17 & 70.73 & 46.18 \bigstrut\\
    \hline
    \multicolumn{1}{|r|}{} & \multicolumn{2}{c|}{Combined} & \multicolumn{2}{c|}{Combined} & \multicolumn{2}{c|}{Combined}\bigstrut\\
    \hline
    Landmark-I \cite{raghavendra2017face} & \multicolumn{1}{c|}{\textbf{99.51}} & \textbf{95.37} & \textbf{96.32} & \textbf{85.43} & \textbf{94.30}  & \textbf{79.25} \bigstrut\\
    \hline
    Landmark-II \cite{UBO_Morphing_Tool} & 90.16 & 71.17 & 90.59 & 66.67 & 83.50  & 57.38 \bigstrut\\
    \hline
    StyleGAN \cite{MorphStyleGAN2020} & 55.06 & 29.39 & 36.36 & 14.83 & 35.62 & 14.28 \bigstrut\\
    \hline
    MIPGAN-I & 63.22 & 35.73 & 40.46 & 28.71 & 61.66 & 34.14 \bigstrut\\
    \hline
    MIPGAN-II & 57.47 & 31.45 & 51.72 & 23.54 & 54.94 & 27.46 \bigstrut\\
    \hline
    \end{tabular}%
    }
  \label{tab:vulnerability-report-Neurotec}%
\end{table*}%

\begin{table*}[htbp]
  \centering
    \caption{Quantitative evaluation of vulnerability of LCNN-29 \cite{wu2018light} FRS from various morph generation approaches. Note that, since FNMR = $0$ $@$ FMR = $0.1\%$ for LCNN-29 \cite{wu2018light} following Eq. \ref{Eqa:RMMR_MMPMR} and \ref{Eqa:RMMR_FMMPMR}, the value of RMMR is equal to MMPMR/FMMPMR. Therefore, we have not entered RMMR separately in the Table above.}
    \resizebox{1\linewidth}{!}{
    \begin{tabular}{|c|c|c|c|c|c|c|}
    \hline
    \cline{2-7}    \multirow{3}[6]{*}{\textbf{Morph generation type\newline{}\newline{}}} & MMPMR/RMMR(\%) & FMMPMR/RMMR(\%) & MMPMR/RMMR(\%) & FMMPMR/RMMR(\%) & MMPMR/RMMR(\%) & FMMPMR/RMMR(\%) \bigstrut\\
\cline{2-7}    \multicolumn{1}{|c|}{} & \multicolumn{2}{c|}{\textbf{Digital}} & \multicolumn{2}{c|}{\textbf{Print-Scan}} & \multicolumn{2}{c|}{\textbf{Print-Scan with compression}} \bigstrut\\
\cline{2-7}    \multicolumn{1}{|c|}{} & \multicolumn{2}{c|}{Male} & \multicolumn{2}{c|}{Male} & \multicolumn{2}{c|}{Male} \bigstrut\\
    \hline
   Landmark-I \cite{raghavendra2017face} & \multicolumn{1}{c|}{96.63} & 89.28 & 95.25 & 89.36 & 94.80 & 88.62 \bigstrut\\
    \hline
    Landmark-II \cite{UBO_Morphing_Tool} & 75.09 & 60.72 & 74.64 & 57.81 & 82.43 & 68.32 \bigstrut\\
    \hline
    StyleGAN \cite{MorphStyleGAN2020} & 83.12 & 66.44 & 85.20 & 69.54 & 84.85 & 68.88 \bigstrut\\
    \hline
    MIPGAN-I & 95.13 & 86.35 & 94.04 & 84.39 & 94.09 & 84.30 \bigstrut\\
    \hline
    MIPGAN-II & 94.93 & 85.14 & 93.94 & 83.14 & 93.75 & 82.63 \bigstrut\\
    \hline
    \multicolumn{1}{|r|}{} & \multicolumn{2}{c|}{Female} & \multicolumn{2}{c|}{Female} & \multicolumn{2}{c|}{Female} \bigstrut\\
    \hline
    Landmark-I \cite{raghavendra2017face} & \multicolumn{1}{c|}{99.16} & 95.00 & 98.75 & 94.26 & 98.96 & 94.49 \bigstrut\\
    \hline
    Landmark-II \cite{UBO_Morphing_Tool} & 92.04 & 82.28 & 94.69 & 82.85 & 95.92 & 86.98 \bigstrut\\
    \hline
    StyleGAN \cite{MorphStyleGAN2020} & 93.33 & 80.08 & 92.92 & 83.06 & 92.92 & 82.76 \bigstrut\\
    \hline
    MIPGAN-I & 97.76 & 92.27 & 96.94 & 91.59 & 96.94 & 91.44 \bigstrut\\
    \hline
    MIPGAN-II & 95.71 & 90.72 & 95.31 & 89.85 & 95.71 & 89.58 \bigstrut\\
    \hline
    \multicolumn{1}{|r|}{} & \multicolumn{2}{c|}{Combined} & \multicolumn{2}{c|}{Combined} & \multicolumn{2}{c|}{Combined}\bigstrut\\
    \hline
    Landmark-I \cite{raghavendra2017face} & \multicolumn{1}{c|}{\textbf{97.16}} & \textbf{90.19} & \textbf{95.96} & \textbf{90.14} & \textbf{95.64} & \textbf{89.55}\bigstrut\\
    \hline
    Landmark-II \cite{UBO_Morphing_Tool} & 78.42 & 64.20 & 78.58 & 61.85 & 85.11 & 71.36 \bigstrut\\
    \hline
    StyleGAN \cite{MorphStyleGAN2020} & 85.12 & 68.61 & 86.72 & 71.69 & 86.44 & 71.09 \bigstrut\\
    \hline
    MIPGAN-I & 95.68 & 87.30 & 94.64 & 85.55 & 94.68 & 85.45 \bigstrut\\
    \hline
    MIPGAN-II & 95.12 & 86.05 & 94.25 & 84.23 & 94.17 & 83.75 \bigstrut\\
    \hline
    \end{tabular}%
    }
  \label{tab:vulnerability-report-LCNN29}%
\end{table*}%

\subsection{Vulnerability Analysis}

This section presents the vulnerability analysis of the proposed morphed face generation techniques to quantify the impact of our efficient attacks on FRS. We quantify the attack success for five different FRS including two Commercial-off-the-Shelf (COTS) FRS and three deep-learning-based open-source FRS. The COTS FRS include the Cognitec FRS (Version 9.4.2) \cite{cognitecfrssdk}  \footnote{Outcome not necessarily constitutes the best the algorithm can do} and Neurotechnology (Version 10) \cite{Neurotech} and the set of open-source FRS includes Arcface \cite{deng2019arcface}, VGGFace \cite{cao2018vggface2} and LCNN-29 \cite{wu2018light} . The operational threshold for all 5 FRS is set at False Match Rate (FMR) of $0.1\%$ following the guidelines of Frontex \cite{EU-Frontex-BestPracticeABC-2015}.

The vulnerability is assessed using two metrics Mated Morphed Presentation Match Rate (MMPMR) \cite{scherhag2017biometric} and Fully Mated Morphed Presentation Match Rate (FMMPMR) \cite{MorphStyleGAN2020} based on the threshold provided by Cognitec FRS. For a given morph image $M_{I_{1,2}}$ obtained using two subjects, we compute the vulnerability by enrolling $M_{I_{1,2}}$ and verifying it against probe images from the corresponding contributing subjects $I_{1}$ and $I_{2}$. The obtained comparison scores $S_1$ and $S_2$ for both probe images $I_{1}$ and $I_{2}$ against the morphed image $M_{I_{1,2}}$ indicates the threat to FRS, if and only if both $S_1$ and $S_2$ cross the actual verification threshold at FMR = 0.1$\%$.  The corresponding metric \textsl{FMMPMR} \cite{MorphStyleGAN2020} \cite{Venkatesh_2020_IJCB} is therefore computed as: 
\begin{dmath}
FMMPMR = \frac{1}{P} \sum_{M,P}^{} {(S1_{M}^{P} > \tau) \&\& (S2_{M}^{P} > \tau) \\ \ldots \&\& (Sk_{M}^{P} > \tau)},
\label{Eqa:FMMPMR}
\end{dmath}
where $P = {1, 2, \ldots, p}$ represent the number of attempts made by presenting all probe images of the contributing subjects against the $M^{th}$ morphed image,  $K = {1, 2, \ldots, k}$ represents the number of composite image constitute to generate the morphed image (in our case  $K=2$), $Sk_{M}^{P}$ represents the comparison score of the $K^{th}$ contributing subject obtained with $P^{th}$ attempt corresponding to $M^{th}$ morphed image and $\tau$ represents the threshold value corresponding to FMR = 0.1$\%$. 
When compared to MMPMR, the FMMPMR will consider both pair-wise comparison of contributory subjects and the number of attempts. In order to also establish the relationship with respect to earlier metrics, we also report the vulnerability using MMPMR \cite{scherhag2017biometric}. 

Further, to effectively analyse the vulnerability, we also present the  results using  Relative Morph Match Rate (RMMR)  defined as follows \cite{Scherhag-MorphingAttacks-MorphingTechniques-BIOSIG-2017}:

\begin{equation}
\begin{aligned}
RMMR(\tau)_{MMPMR} =& 1+(MMPMR(\tau))\\
&-[1-FNMR(\tau)]
\end{aligned}
\label{Eqa:RMMR_MMPMR}
\end{equation}
\begin{equation}
\begin{aligned}
RMMR(\tau)_{FMMPMR} =& 1 + (FMMPMR(\tau))\\
&-[1-FNMR(\tau)]
\end{aligned}
\label{Eqa:RMMR_FMMPMR}
\end{equation}

Where, FNMR indicates the False Reject Rate (FNMR) of the FRS under consideration obtained at the threshold $\tau$. In this work, $\tau$ represents the value corresponding to FMR = 0.1$\%$. Since we have evaluated 5 different FRS systems, we have computed FNMR corresponding to these FRS to calculate the RMMR.  Note that, in Equation \ref{Eqa:RMMR_MMPMR} and \ref{Eqa:RMMR_FMMPMR} if FNMR = 0 then RMMR corresponds to MMPMR/FMMPMR.

The obtained success rate, or alternatively the vulnerability of FRS is provided in Table~\ref{tab:vulnerability-report}, \ref{tab:vulnerability-report-VGGFace2}, \ref{tab:vulnerability-report-Arcface}, \ref{tab:vulnerability-report-Neurotec} and \ref{tab:vulnerability-report-LCNN29} corresponding to to Cognitec \cite{cognitecfrssdk},  VGGFace \cite{cao2018vggface2}, Arcface \cite{deng2019arcface}, Neurotechnology (Version 10) \cite{Neurotech} and LCNN-29 \cite{wu2018light} respectively. The vulnerability analysis is carried out on 5 different morph generation methods that include facial landmarks (Landmarks-I) with image smoothing as the post-processing operation \cite{raghavendra2017face}, Facial landmarks (Landmarks-II) with automatic image retouching and colour equalisation \cite{UBO_Morphing_Tool}, existing GAN based face morphing method based on StyleGAN \cite{MorphStyleGAN2020} and proposed MIPGAN variants (MIPGAN-I and MIPGAN-II).  Based on the obtained results, the following are the concrete observations: 
\begin{itemize}
\item The FNMR corresponding to five different FRS is equal to $0$. Therefore, the value of the RMMR  is equal to MMPMR or FMMPMR. This indicates that the FRS systems are accurate on our face datasets employed in this work. 
\item  Among the five FRS, the highest vulnerability is noted for Arcface \cite{deng2019arcface}, which is vulnerable to all five kinds of face morphing attack methods. 
\item Among COTS FRS, the Cognitec FRS indicates a higher vulnerability on all five types of face morphing attack methods compared to Neurotechnology  FRS. 
\item Among five different morph generation methods, Landmark-I indicates the highest vulnerability on all five other FRS. 
\item The proposed face morphing methods MIPGAN-I and MIPGAN-II  consistently indicate the highest vulnerability, when compared to the existing method based on  StyleGAN \cite{MorphStyleGAN2020}. This indicates the high quality of morphs generated using the proposed MIPGAN-I and MIPGAN-II methods. 
\item The proposed MIPGAN-I and MIPGAN-II methods also indicate a higher vulnerability than the Landmark-II technique for morph generation with four different FRS. 
\item Among the two different metrics (MMPMR and FMMPMR), the proposed FMMPMR indicates a lower vulnerability than MMPMR consistently as FMMPMR imposes a strict selection of attack images, unlike MMPMR. 
\item  MIPGAN-I based morphed images show a marginally better performance in attacking FRS than images generated by MIPGAN-II.
\end{itemize}

\subsection{Perceptual Image Quality Analysis}
\begin{figure}[htp]
	\centering
	\includegraphics[width=1\linewidth]{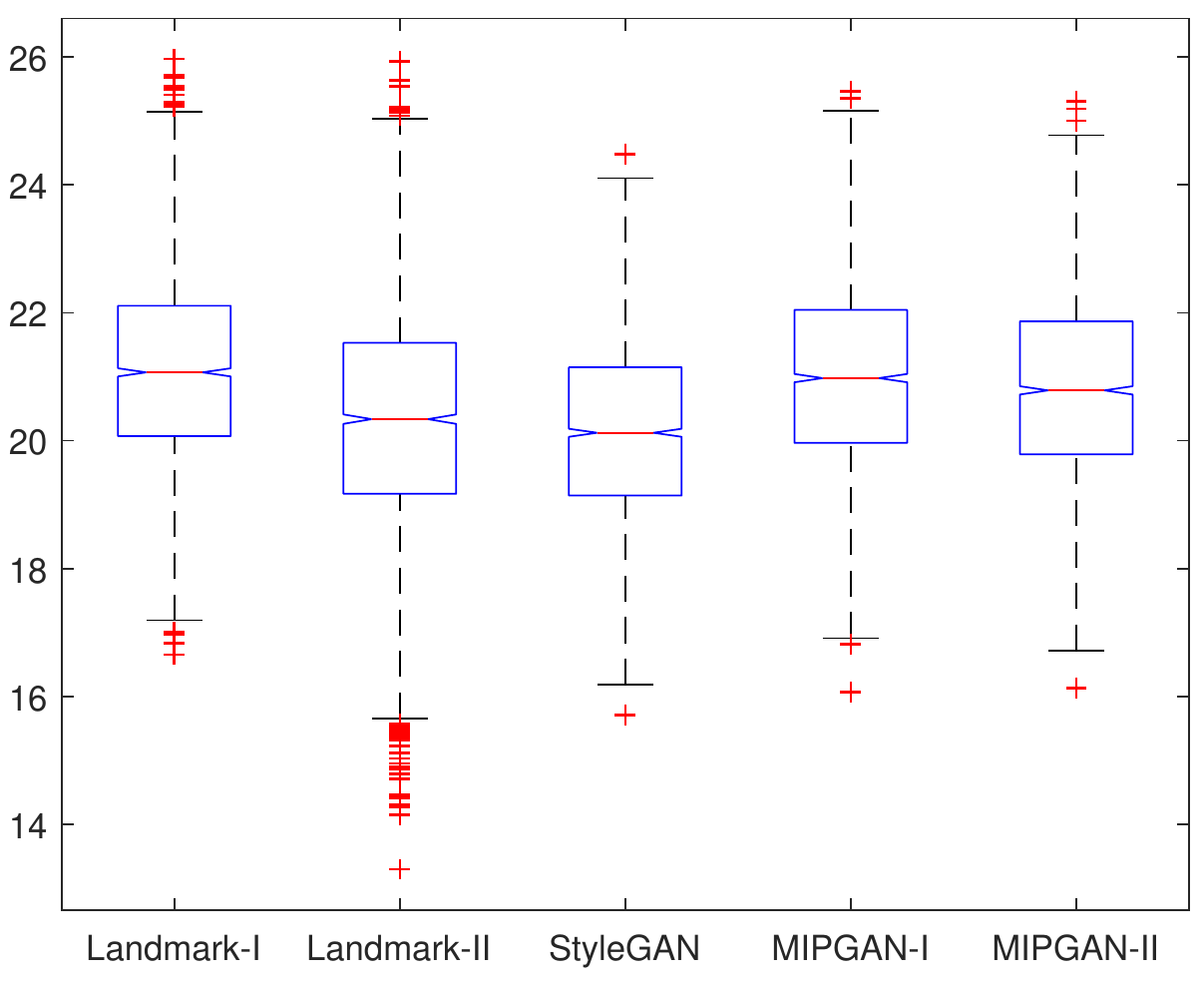}
	\caption{Box plots of PSNR values computed from different face morph generation methods (digital version)}
	\label{fig:PSNRBOx}
\end{figure}

\begin{figure}[htp]
	\centering
	\includegraphics[width=1\linewidth]{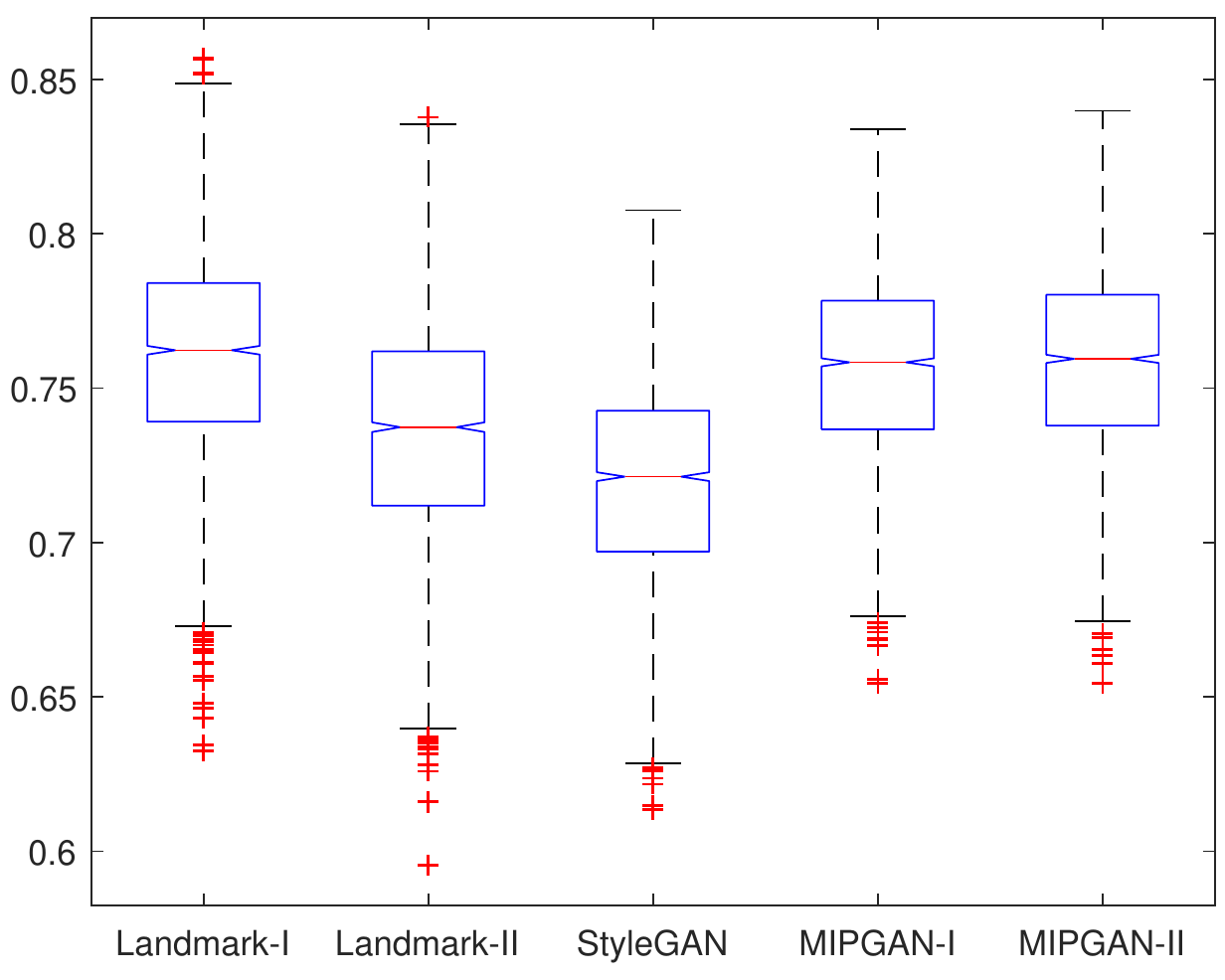}
	\caption{Box plots of SSIM values computed from different face morph generation methods (digital version)}
	\label{fig:SSIMBOx}
\end{figure}

This section presents quantitative results of the proposed morphed image generation techniques using the perceptual image quality metrics PSNR and SSIM. Both of these metrics are computed based on the reference image. Morphed face images are generated based on parent face images from two contributory data subjects. Therefore, we used the parent face images from both contributory data subjects as the reference image against which the given morphed image is assessed and we average the obtained image quality scores for both parent images.  Table \ref{tab:PSNRnSSIM} indicates the quantitative results of both PSNR and SSIM on four different types of face morph generation mechanism in the digital format. Based on the obtained results, it can be observed that:
\begin{itemize}
\item There is little deviation in the perceptual image quality metrics computed on all four different types of face morph generation mechanisms. 
\item The proposed MIPGAN-I and MIPGAN-II methods indicate a slightly better image quality when compared to the StyleGAN \cite{MorphStyleGAN2020} based face morphing method. 
\item The proposed MIPGAN-I and facial landmarks-based methods \cite{UBO_Morphing_Tool} indicate a similar image quality.  
\item Figure \ref{fig:PSNRBOx} and \ref{fig:SSIMBOx} indicate the box plots of the PSNR and SSIM quality scores. These results further indicate that the perceptual quality of the proposed MIPGAN-I and MIPGAN-II is superior to the existing state-of-the-art method based on StyleGAN \cite{MorphStyleGAN2020}.  
\end{itemize}
\begin{table}[htbp]
  \centering
  \caption{Morph image quality analysis using PSNR and SSIM with 95\% confidence interval}
  \resizebox{0.9\linewidth}{!}{
    \begin{tabular}{|c|c|c|}
    \hline
    \textbf{Morph generation Methods} & \multicolumn{1}{c|}{\textbf{PSNR}} & \multicolumn{1}{c|}{\textbf{SSIM}} \bigstrut\\
    \hline
    Landmark-I \cite{raghavendra2017face}  & 21.1111$\pm$ 0.0415
 & 0.7609$\pm$0.0009 \bigstrut\\
    \hline
    Landmark-II \cite{UBO_Morphing_Tool}  &    20.2737$\pm$0.0523
 & 0.7363$\pm$0.0010 \bigstrut\\
    \hline
    StyleGAN \cite{MorphStyleGAN2020} & 20.1347$\pm$0.0383 & 0.7199$\pm$0.0008 \bigstrut\\
    \hline
    MIPGAN-I & 21.0133$\pm$0.0409
 & 0.7573$\pm$0.0008 \bigstrut\\
    \hline
    MIPGAN-II & 20.8306$\pm$0.0409 & 0.7586$\pm$0.0008 \bigstrut\\
    \hline
    \end{tabular}%
    }
  \label{tab:PSNRnSSIM}%
\end{table}%

\subsection{Human Observer Analysis}


\begin{figure*}[htbp]
\centering
\subfigure[]{
\centering
\includegraphics[width=0.3\linewidth]{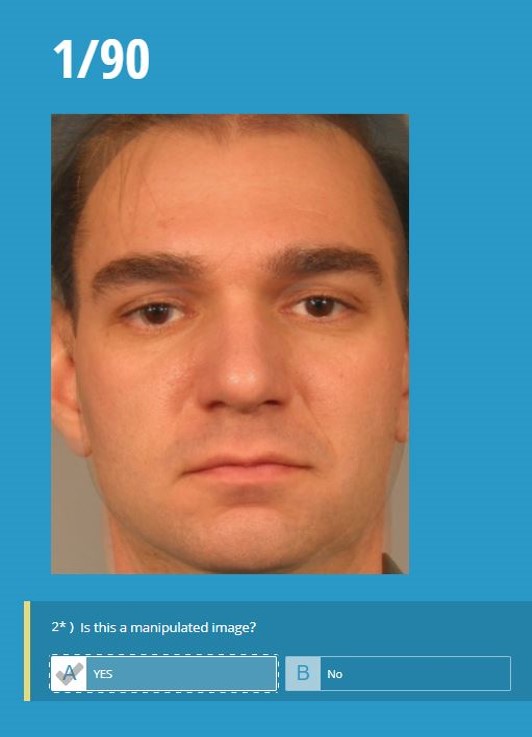}
}%
\subfigure[]{
\centering
\includegraphics[width=0.6\linewidth]{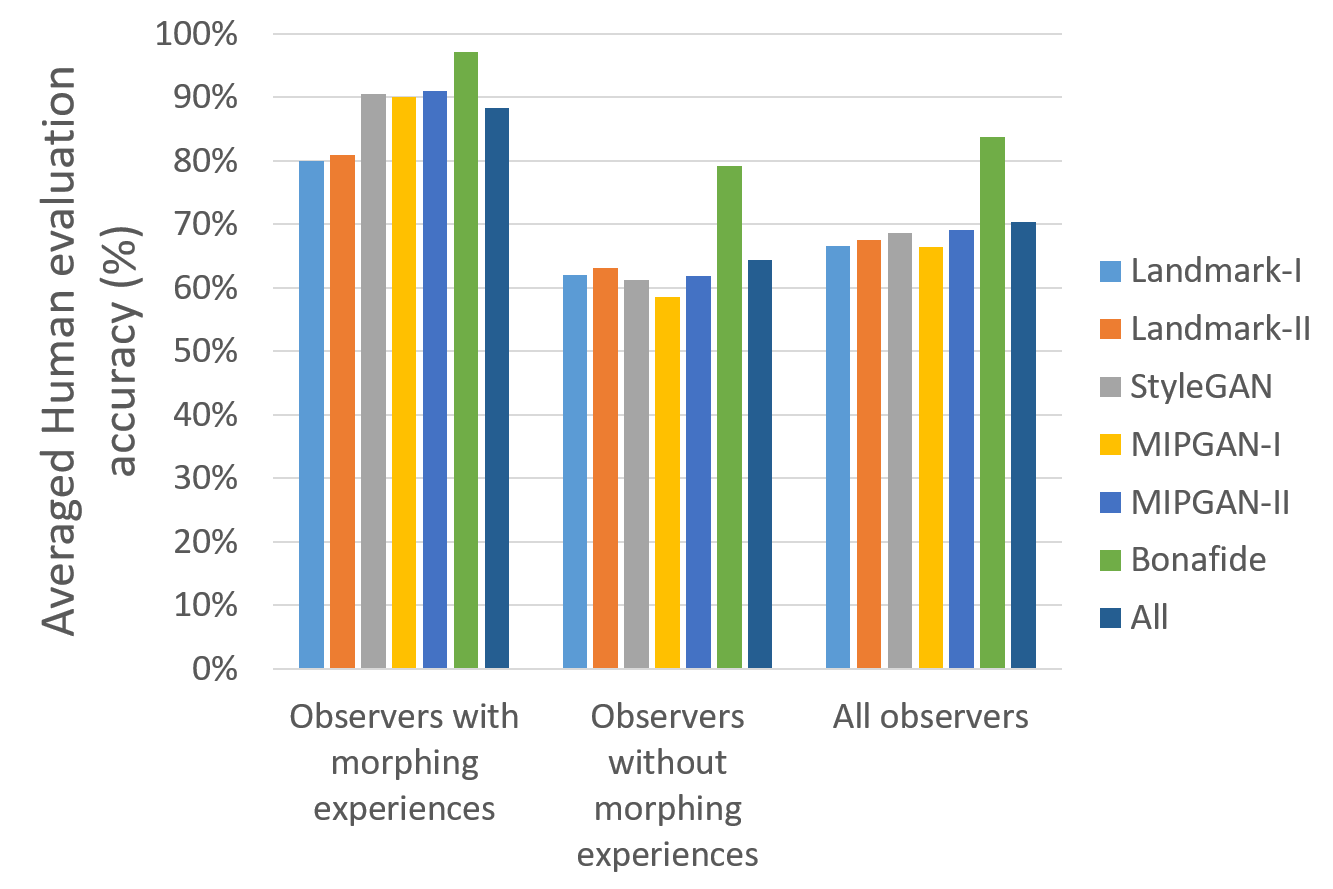}
}%
\centering
\caption{(a)Example of screen shot used for human observer study (b) Quantitative results}
\label{fig:HumanObservers}
\end{figure*}

\begin{figure*}[htbp]
\centering
\subfigure[]{
\centering
\includegraphics[width=0.4\linewidth]{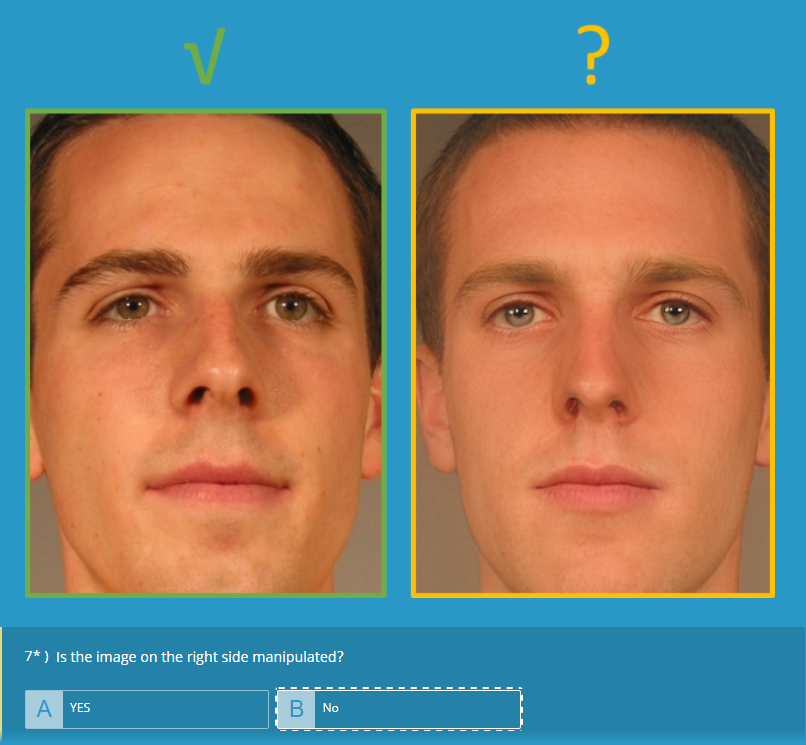}
}%
\subfigure[]{
\centering
\includegraphics[width=0.6\linewidth]{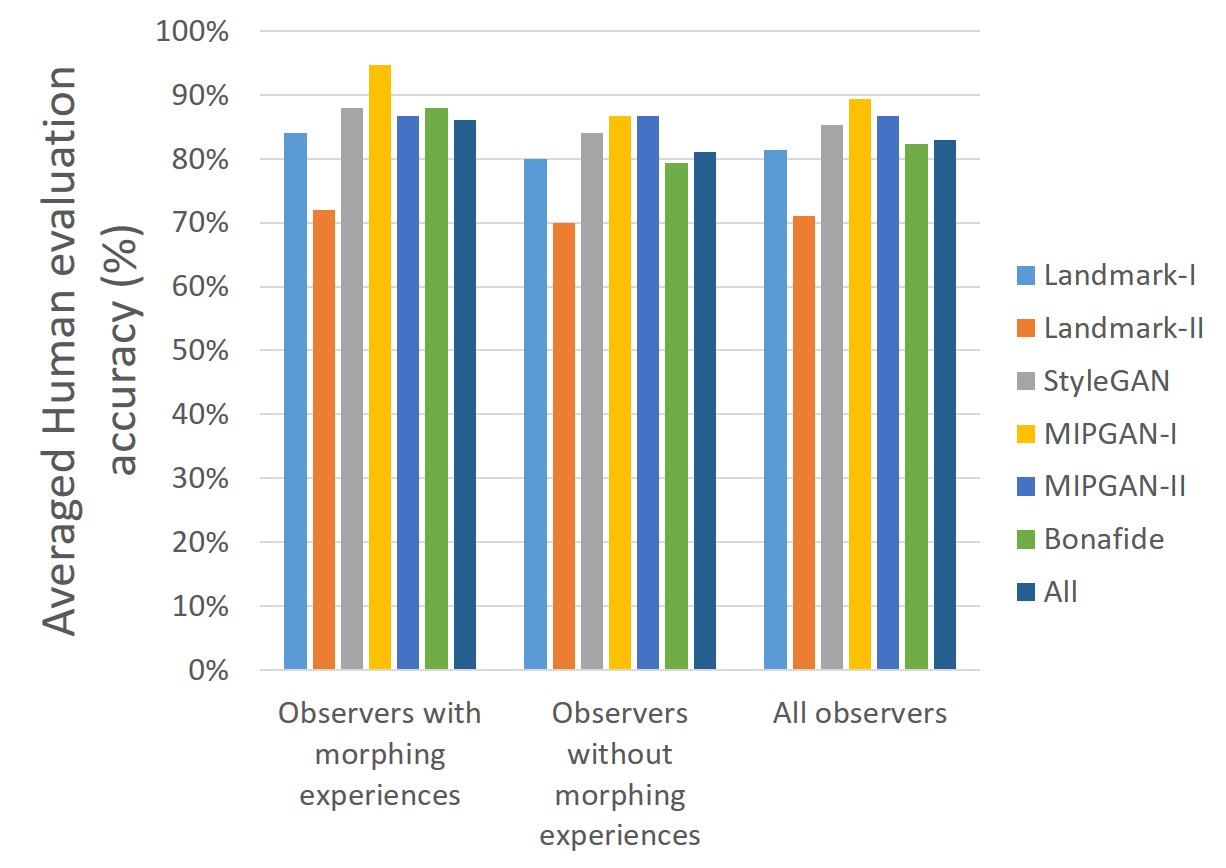}
}%
\centering
\caption{(a)Example of screen shot used for differential human observer study (b) Quantitative results}
\label{fig:HumanObserversD}
\end{figure*}
\begin{table*}[htbp]
  \centering
  \caption{Vulnerability - Ablation study on the proposed loss function. Here, \ding{51} indicates the selected and \ding{53} indicates the not selected loss function in the ablation study}
  \resizebox{1\linewidth}{!}{
    \begin{tabular}{|p{5.4em}|p{6.135em}|p{6.4em}|p{6.765em}|c|c|c|c|c|c|c|c|}
    \hline
    \renewcommand{\multirowsetup}{\centering}
    \multirow{3}[6]{*}{\textbf{$Loss_{ID-Diff}$}} & \multirow{3}[6]{*}{\textbf{$Loss_{Identity}$}} & \multirow{3}[6]{*}{\textbf{$Loss_{MS-SSIM}$}} & \multirow{3}[6]{*}{\textbf{$Loss_{Perceptual}$}} & \multicolumn{4}{c|}{\textbf{MIPGAN-I}} & \multicolumn{4}{c|}{\textbf{MIPGAN-II}} \bigstrut\\
\cline{5-12}   \multicolumn{1}{|c|}{} & \multicolumn{1}{c|}{} & \multicolumn{1}{c|}{} & \multicolumn{1}{c|}{} & \multicolumn{2}{c|}{FMMPMR} & \multicolumn{2}{c|}{MMPMR} & \multicolumn{2}{c|}{FMMPMR} & \multicolumn{2}{c|}{MMPMR} \bigstrut\\
\cline{5-12}    \multicolumn{1}{|c|}{} & \multicolumn{1}{c|}{} & \multicolumn{1}{c|}{} & \multicolumn{1}{c|}{} & \multicolumn{1}{c|}{Cognitec} & \multicolumn{1}{c|}{ArcFace} & \multicolumn{1}{c|}{Cognitec} & \multicolumn{1}{c|}{ArcFace} & \multicolumn{1}{c|}{Cognitec} & \multicolumn{1}{c|}{ArcFace} & \multicolumn{1}{c|}{Cognitec} & \multicolumn{1}{c|}{ArcFace} \bigstrut\\
    \hline
    \centering\ding{53} & \centering\ding{51}    & \centering\ding{51}     & \centering\ding{51}     & 81.82 & 75.87 & 90.69 & 93.47  & 77.83 & 71.98 & 90.1  & 91.18 \bigstrut\\
    \hline
    \centering\ding{51}     & \centering\ding{53} & \centering\ding{51}     & \centering\ding{51}    & 78.07    & 62.15 & 89.17 & 83.77 & 78.39 & 64.51 & 90.04 & 82.54 \bigstrut\\
    \hline
   \centering\ding{51}     & \centering\ding{51}    & \centering\ding{53} & \centering\ding{51}   & 80.82 & 73.33 & 91.81 & 92.66 & 78.73 & 71.79 & 89.58 & 90.55 \bigstrut\\
    \hline
    \centering\ding{51}     & \centering\ding{51}    & \centering\ding{51}     & \centering\ding{53}    & 21.37 & 47.85 & 44.18  & 71.95 & 11.92 & 33.12 & 29.47 & 59.56 \bigstrut\\
    \hline
   \centering\ding{51}    & \centering\ding{51}     & \centering\ding{51}     & \centering\ding{51}  & \textbf{84.65} & \textbf{85.94} & \textbf{94.36} & \textbf{94.45} & \textbf{81.59} & \textbf{86.24} & \textbf{92.93} & \textbf{94.21} \bigstrut\\
    \hline
    \end{tabular}%
    }
  \label{tab:Abalation}%
\end{table*}%

In this section, we discuss the quantitative detection performance of human observations regarding morphed face images, which are generated using MIPGAN-I and MIPGAN-II. To this extent, we have designed and developed a web-portal to evaluate the human morph detection performance reflecting both single image-based morphing attack detection scenario (S-MAD) and differential morphing attack detection scenario (D-MAD). We have used only digital samples of both bona fide and morphed face images as the proposed  MIPGAN is used to generate the images in the digital domain.  Figure \ref{fig:HumanObservers} (a) shows the screenshot of the web-portal for S-MAD in which the human observer needs to decide whether the displayed image is a morphed face image or a bona fide image by looking at one single image at a time. Correspondingly, Figure \ref{fig:HumanObserversD} (a) presents the screenshot for D-MAD experiment where the observer needs to detect whether the unknown image is morphed given a trusted bona fide image as a reference. We have selected a total of 90 images where 15 images are from each group corresponding to bona fide, two different types of facial landmarks based morphing such as Landmarks-I \cite{raghavendra2017face} and Landmarks-II \cite{UBO_Morphing_Tool}, StyleGAN \cite{MorphStyleGAN2020} based face morphing, MIPGAN-I and MIPGAN-II based face morphing. To make the testing robust, all 90 chosen images correspond to unique data subjects and there is no repetition of data subjects. To avoid gender bias by participants, we have selected a near equal distribution of male and female data subjects in each group. We have chosen 90 images considering the time constraints required to assess these images for human observers. It was important that observers do not loose focus while conducting the detection experiments. 

Figure \ref{fig:HumanObservers} (b) shows the quantitative results of S-MAD obtained from 56 human observers, including 14 experienced and 42 inexperienced observers. The experienced observers' group consists of researchers working in face morphing attack detection and as ID expert's in border control, while the non-experienced group consists of students and other computer science professionals. As noticed from the Figure \ref{fig:HumanObservers} (b) following are the main observations: 
\begin{itemize}
 \item Detection performance of the bona fide images indicates better detection performance by both experienced and non-experienced group when compared to the morphed face image.  The experienced group indicates the detection performance with an accuracy of  97.14\%, while the non-experienced group indicates the detection performance with an accuracy of 79.21\%.    
 \item Human observers with experience in face morphing demonstrate higher detection accuracy on four different face morph generation mechanisms than the inexperienced group.  
 \item Among the four different morphing types, the experienced group indicates that the detection of the landmarks-based morphing is challenging compared to other morphing mechanisms (deep learning-based). 
 \item Human observers with no experience in face morphing are marginally good in detecting the landmarks-based face morph images compared to other types of face morphing techniques. MIPGAN-I exhibits more challenging morph images to detect as compared to other morph generation methods. 
\item Based on the obtained results, it can be noted that the human observers with good experience in face morphing can detect morphed images with an accuracy of 88.25\% while the human observer with no knowledge of face morphing shows the challenge to detect the morphed face images with a detection accuracy of 64.31\%. 
\item The overall results from 56 human observers indicate that detecting morphed face images is challenging. Further, it is also interesting to note that detecting different face morphing types is also challenging. 
\end{itemize}

For the quantitative results of D-MAD, 5 experienced observers and 10 inexperienced observers have participated. As shown in Figure \ref{fig:HumanObserversD} (b), the following observations are illustrated:
\begin{itemize}
    \item In the scenario of D-MAD, the group with relevant experiences achieved an overall 86\% accuracy, which is better than 81\% for the inexperienced group. However, this difference is much less than the difference in S-MAD, which means that the reference image can help inexperienced observers to identify the morphs. 
    \item Morphs generated by Landmark-II present a significant challenge as compared to  other morph generation mechanisms in D-MAD. This may be attributed to a more natural skin texture appearance (comparing with GAN-based mechanisms) and fewer artefacts (comparing with Landmark-I) and observers focusing less on its minor artefacts in the pairwise comparison.
    \item It is also interesting to see that the performances of experienced observers on detecting Landmark-II (80.95\% and 72.00\%), StyleGAN (90.48\% and 88.00\%), MIPGAN-II (90.95\% and 86.67\%), and bona fide images (90\% and 88.00\%) are lower than their performance in S-MAD. We believe this is because experienced observers do not pay critical attention to tolerable difference between the trusted reference image and the unknown comparison image. 
\end{itemize}

\subsection{Ablation Study}
\label{sec:Ablation}
In order to measure the impact of the loss functions in the proposed approach, we conduct an extensive ablation study. The proposed loss function combines four different entities such as: perceptual loss ($Loss_{Perceptual}$), identity loss ($Loss_{Identity}$), identity difference ($Loss_{ID-Diff}$) and  Multi-Scale  Structural  Similarity  (MS-SSIM)  loss ($Loss_{MS-SSIM}$). The main contribution of our work is to use identity information, which can be considered as a specific high-level feature, to measure the loss. However, high-level features also mean that it is hard for the gradient descent algorithm to ensure a good convergence during the optimisation process. Therefore, we have introduced the perceptual loss that can measure relatively low-level features in addition to MS-SSIM and identity difference loss to effectively control the optimisation process to generate a high-quality morphed image. We perform the ablation study by discarding each term in the loss function iteratively. We benchmark the vulnerability using COTS FRS (Cognitec FRS (Version 9.4.2)) and the open-source ArcFace FRS, as the proposed approach is dedicated to generating high-quality morphed images.

\begin{figure}[htp]
	\centering
	\includegraphics[width=1\linewidth]{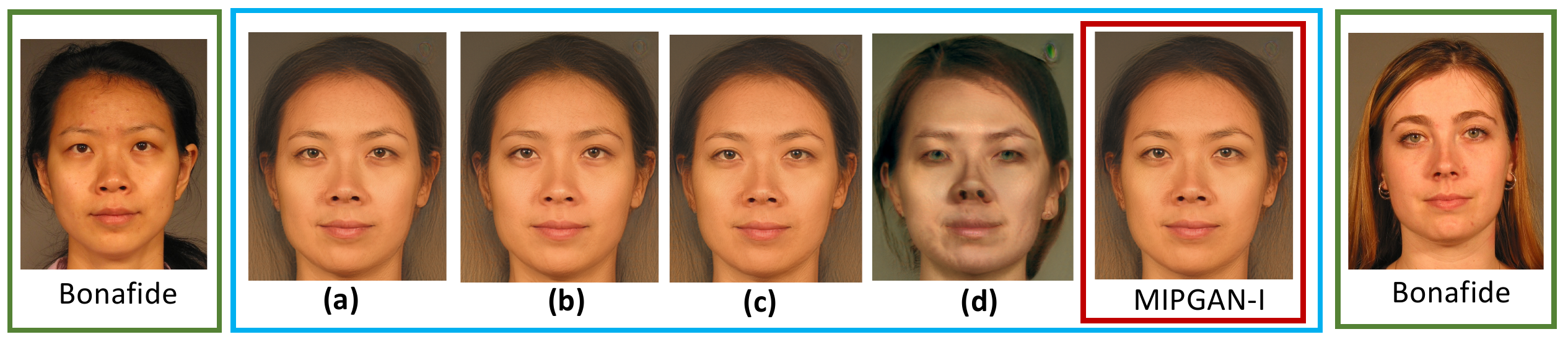}
	\caption{Qualitative results of ablation study using proposed MIPGAN-I (a)$Loss_{ID-Diff}$ (b) $Loss_{Identity}$ (c) $Loss_{MS-SSIM}$ (d) $Loss_{Perceptual}$}
	\label{fig:AblationMIPGAN1}
\end{figure}

\begin{figure}[htp]
	\centering
	\includegraphics[width=1\linewidth]{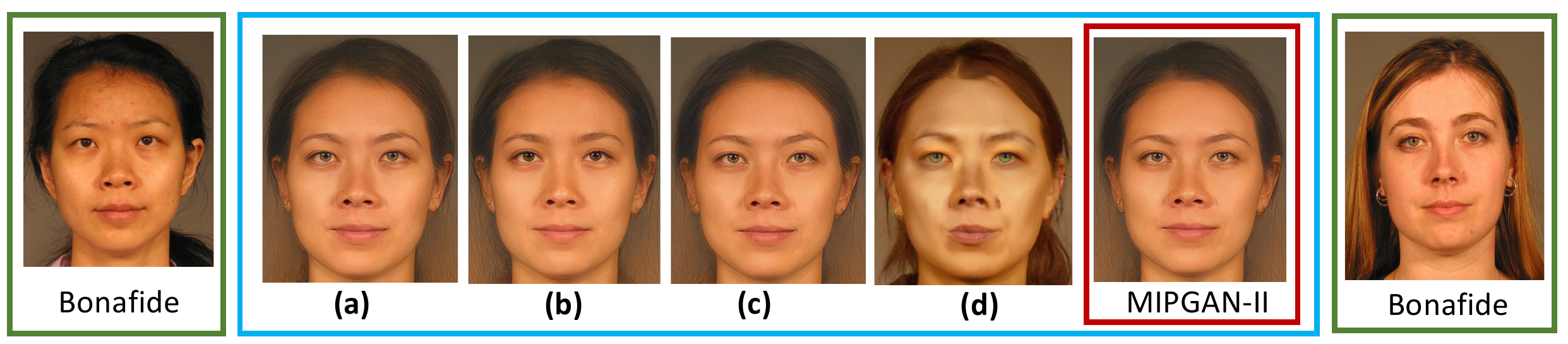}
	\caption{Qualitative results of ablation study using proposed MIPGAN-II (a)$Loss_{ID-Diff}$ (b) $Loss_{Identity}$ (c) $Loss_{MS-SSIM}$ (d) $Loss_{Perceptual}$}
	\label{fig:AblationMIPGAN2}
\end{figure}

Table \ref{tab:Abalation} indicates the quantitative performance of the ablation study using a vulnerability analysis for both the COTS-FRS from Cognitec and for the open-source Arcface FRS with the proposed MIPGAN-I and MIPGAN-II methods. The ablation study is carried out on the digital morphed images generated using both MIPGAN-I and MIPGAN-II Methods. Figure \ref{fig:AblationMIPGAN1} and \ref{fig:AblationMIPGAN2} shows the qualitative performance of the ablation study on both MIPGAN-I and MIPGAN-II, respectively. Based on the obtained results, the following are the main observations: 
\begin{itemize}
 \item Each term in our proposed loss function (see Eq. \ref{eqn:Finalloss})  contributes to posing a greater challenge to a FRS for both proposed MIPGAN-I and MIPGAN-II morph generation frameworks. 
 \item Among the four other loss functions that we have used, the $Loss_{Perceptual}$ is critical in improving the proposed method's performance. Discarding the perceptual loss has resulted in a degrading performance in both qualitative (see Figure \ref{fig:AblationMIPGAN1} (d) and \ref{fig:AblationMIPGAN2} (d)) and quantitative  results. 
 \item The use of identity loss ($Loss_{Identity}$) also indicates the importance of improving the quantitative performance of the proposed method.  
 \item The $Loss_{MS-SSIM}$ also contributes to both qualitative and quantitative improvements of the morphs generated by the proposed method. 
\end{itemize}

\subsection{Hyper-parameters Study}
\label{sec:Hyper}
This section presents both qualitative and quantitative results on the selection of hyper-parameters ($\lambda_1$, $\lambda_2$, $\lambda_3$, and $\lambda_4$) in the proposed loss function employed in both MIPGAN-I and MIPGAN-II.  Based on the ablation study reported in Section \ref{sec:Ablation}, we have noticed that the perceptual loss is the vital component of our loss function (see Eq. \ref{eqn:Finalloss}) and the other three terms can be used as constraints during the optimisation. Therefore, the first step is to study the generated morphed face images' attack strength by increasing and decreasing the value of $\lambda_1$. Among the remaining three terms, we have also noticed from the ablation study that the identity loss ($Loss_{Identity}$) is contributing more towards generating a high-quality morph compared to the other two-loss functions ($ loss_{MS-SSIM}$, $ Loss_{ID-Diff} $). We analyze the importance of identity loss ($Loss_{Identity}$) with respect to the other two loss functions ($Loss_{MS-SSIM}$, $ Loss_{ID-Diff} $) by increasing the value of $\lambda_3$ and/or  $\lambda_3$ and decreasing the value of $\lambda_2$. Further, we have also noticed from the ablation study that the loss functions $ loss_{MS-SSIM}$ and $ Loss_{ID-Diff} $ are less important and numerically very small. Therefore, we did not conduct studies on decreasing the values of $\lambda_3$ and $\lambda_4$.  Altogether, we have tested four different cases of changing the hyper-parameter values to generate the morphed face images. These generated morphed face images are benchmarked against the proposed hyper-parameter values through the vulnerability analysis using both COTS FRS (Cognitec FRS (Version 9.4.2)) and open-source ArcFace FRS.

\begin{figure}[htp]
	\centering
	\includegraphics[width=1\linewidth]{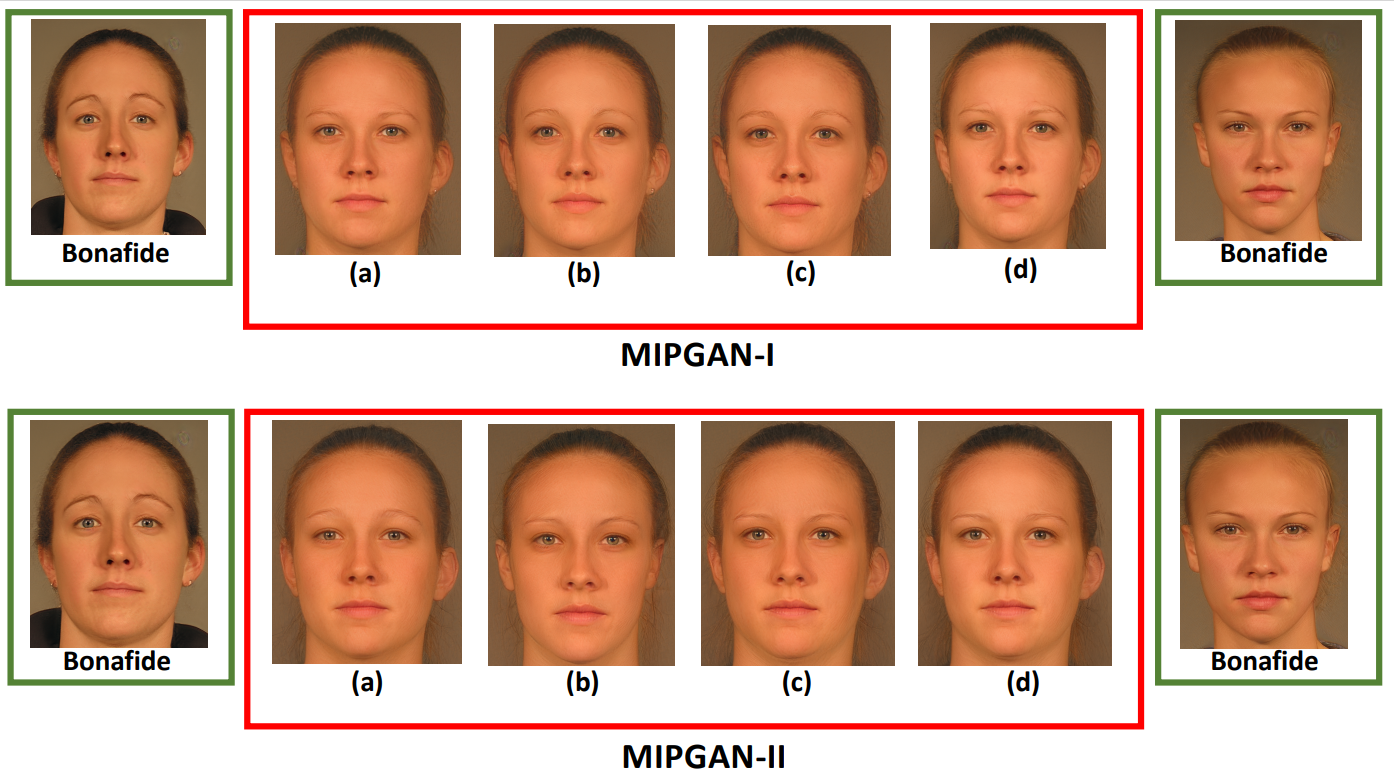}
	\caption{Qualitative results of Hyper-parameters study on both MIPGAN-I and  MIPGAN-II (a)$\lambda_1$ (b) $\lambda_2$ (c) $\lambda_3$ (d) $\lambda_4$}
	\label{fig:HyperParamters}
\end{figure}

\begin{table*}[htbp]
  \centering
  \caption{Quantitative results of hyper-parameters study}
  \resizebox{1\linewidth}{!}{
    \begin{tabular}{|c|c|c|c|c|c|c|c|c|c|}
    \hline
    \multicolumn{1}{|c|}{\multirow{2}[4]{*}{Proposed Morph Generators}} & \multirow{2}[4]{*}{Case-study} & \multicolumn{4}{c|}{Hyper-parameters weights} & \multicolumn{2}{c|}{MMPMR (\%)} & \multicolumn{2}{c|}{FMMPMR (\%)} \bigstrut\\
\cline{3-10}          & \multicolumn{1}{c|}{} & $\lambda_1$    & $\lambda_2$     & $\lambda_3$     & $\lambda_4$    & \multicolumn{1}{c|}{Cognitec} & \multicolumn{1}{c|}{ArcFace} & \multicolumn{1}{c|}{Cognitec} & \multicolumn{1}{c|}{ArcFace} \bigstrut\\
    \hline
    \multicolumn{1}{|c|}{\multirow{5}[10]{*}{MIPGAN -I}} & \centering$1$   & 0.0004 & 10    & 1     & 1     & 93.94 & 93.49 & 84.39 & 75.61 \bigstrut\\
\cline{2-10}          & \centering$2$   & 0.0001 & 10    & 1     & 1     & 92.66 & 91.15 & 79.66 & 72.94 \bigstrut\\
\cline{2-10}          & \centering$3$   & 0.0002 & 1     & 10    & 1     & 94.17 & 91.9  & 84.34 & 75.66 \bigstrut\\
\cline{2-10}          & \centering$4$   & 0.0002 & 1     & 1     & 10    & 83.16 & 82.14 & 67.19 & 59.46 \bigstrut\\
\cline{2-10}          & \centering Proposed weights& 0.0002 & 10    & 1     & 1     & \textbf{94.36} & \textbf{94.45} & 8\textbf{4.65} & \textbf{85.94} \bigstrut\\
    \hline\hline
    \multicolumn{1}{|c|}{\multirow{5}[10]{*}{MIPGAN -II}} & \centering$1$   & 0.0004 & 10    & 1     & 1     & 91.36 & 91.98 & 81.29 & 76.18 \bigstrut\\
\cline{2-10}          & \centering$2$   & 0.0001 & 10    & 1     & 1     & 91.69 & 88.29 & 73.91 & 68.16 \bigstrut\\
\cline{2-10}          & \centering$3$   & 0.0002 & 1     & 10    & 1     & 90.63 & 90.91 & 80.76 & 75.87 \bigstrut\\
\cline{2-10}          & \centering$4$  & 0.0002 & 1     & 1     & 10    & 87.22 & 74.33 & 57.43 & 51.91 \bigstrut\\
\cline{2-10}          & \centering Proposed weights& 0.0002 & 10    & 1     & 1     & \textbf{92.93} & \textbf{94.21} & \textbf{81.59} & \textbf{86.94} \bigstrut\\
    \hline
    \end{tabular}%
    }
  \label{tab:hyperparameterRes}%
\end{table*}%

Table \ref{tab:hyperparameterRes} shows the qualitative performance and Figure \ref{fig:HyperParamters} shows the qualitative performance of the hyper-parameter study. Based on the obtained results, it can be noted that the increase in the value of  $\lambda_1$ and $\lambda_3$  shows comparable results with the proposed weighting schemes. However, based on our empirical study on hyper-parameters, we noted that:  if we set $\lambda_1$ and $\lambda_2$ with equal weights, then, during the optimisation, the generated morph image will soon become roughly similar to both contributing subjects. This will quickly reduce identity loss ($Loss_{Identity}$) to a minimal value and loose its importance in the optimisation.  Hence, we set a larger factor to the identity loss compared with other loss terms measuring high-level features to ensure our most important constraint term is still effective in the later stage of optimisation.  Further, both $\lambda_3$ and $\lambda_4$ can make the optimisation goal more comprehensive but setting a large factor will obstruct the convergence. Especially setting high values to $\lambda_4$ will end up with an image not similar to both subjects. Therefore, the selection of the proposed hyper-parameters confirms the generation of a high-quality morphed image but also aids for effective and comprehensive optimisation.

\subsection{Morphing Attack Detection Potential}
\label{sec:MAD}


\begin{table*}[htbp]
  \centering
  \caption{Quantitative performance of MAD - Training- Landmarks-I \cite{raghavendra2017face}}
  \resizebox{1\linewidth}{!}{
    \begin{tabular}{|c|c|c|c|c|c|c|c|c|c|c|c|}
    \hline
    \multicolumn{1}{|c|}{\multirow{3}[6]{*}{\textbf{Morph Generation Type: Training}}} & \multicolumn{1}{c|}{\multirow{3}[6]{*}{\textbf{Morph Generation Type: Testing}}} & \multirow{3}[6]{*}{\textbf{MAD Algorithms}} & \multicolumn{3}{c|}{\textbf{Digital}} & \multicolumn{3}{c|}{\textbf{Print-scan}} & \multicolumn{3}{c|}{\textbf{Print-scan with compression}} \bigstrut\\
\cline{4-12}          &       & \multicolumn{1}{c|}{} & \multicolumn{1}{c|}{\multirow{2}[4]{*}{D-EER(\%)}} & \multicolumn{2}{p{8.6em}|}{BPCER @ APCER =  } & \multicolumn{1}{c|}{\multirow{2}[4]{*}{D-EER(\%)}} & \multicolumn{2}{p{8.6em}|}{BPCER @ APCER =  } & \multicolumn{1}{c|}{\multirow{2}[4]{*}{D-EER(\%)}} & \multicolumn{2}{p{8.6em}|}{BPCER @ APCER =  } \bigstrut\\
\cline{5-6}\cline{8-9}\cline{11-12}          &       & \multicolumn{1}{r|}{} &       & 5\%   & 10\%  &       & 5\%   & 10\%  &       & 5\%   & 10\% \bigstrut\\
    \hline
    \multicolumn{1}{|c|}{\multirow{10}[20]{*}{\textbf{Landmarks-I \cite{raghavendra2017face}}}} & \cellcolor{mygray}
     & \cellcolor{mygray}Ensemble Features \cite{EnsembleFeatures_2020} & \cellcolor{mygray}0     & \cellcolor{mygray}0     & \cellcolor{mygray}0     & \cellcolor{mygray}2.35  & \cellcolor{mygray}1.45  & \cellcolor{mygray}0.96  & \cellcolor{mygray}2.58  & \cellcolor{mygray}1.71  & \cellcolor{mygray}1.54 \bigstrut\\
\cline{3-12}        &  \multicolumn{1}{c|}{\multirow{-2}{*}{\cellcolor{mygray}Landmarks-I \cite{raghavendra2017face}}}     & \cellcolor{mygray}Hybrid Features \cite{RagISBA2019} & \cellcolor{mygray}0.16  & \cellcolor{mygray}0     & \cellcolor{mygray}0     & \cellcolor{mygray}1.85  & \cellcolor{mygray}0.85  & \cellcolor{mygray}0.34  & \cellcolor{mygray}2.25  & \cellcolor{mygray}1.12  & \cellcolor{mygray}0.51\bigstrut\\
\cline{2-12}          & \multicolumn{1}{c|}{\multirow{2}[4]{*}{Landmarks-II \cite{UBO_Morphing_Tool}}} & Ensemble Features \cite{EnsembleFeatures_2020} & 49.55 & 92.22 & 88.85 & 41.93 & 81.45 & 76.25 & 42.15 & 83.88 & 77.64 \bigstrut\\
\cline{3-12}          &       & Hybrid Features \cite{RagISBA2019} & 49.16 & 99.31 & 97.59 & 44.17 & 86.48 & 80.24 & 46.49 & 88.38 & 81.95 \bigstrut\\
\cline{2-12}          & \multicolumn{1}{c|}{\multirow{2}[4]{*}{StyleGAN \cite{MorphStyleGAN2020}}} & Ensemble Features \cite{EnsembleFeatures_2020} & 0.22  & 0     & 0     & 13.36 & 27.44 & 16.46 & 14.77 & 27.27 & 19.38 \bigstrut\\
\cline{3-12}          &       & Hybrid Features \cite{RagISBA2019} & 0.16  & 0     & 0     & 44.96 & 83.7  & 75.47 & 9.44  & 14.57 & 9.14 \bigstrut\\
\cline{2-12}          & \multicolumn{1}{c|}{\multirow{2}[4]{*}{MIPGAN-I}} & Ensemble Features \cite{EnsembleFeatures_2020}& 39.16 & 73.14 & 65.35 & 9.45  & 14.57 & 8.74  & 8.95  & 15.26 & 9.26 \bigstrut\\
\cline{3-12}          &       & Hybrid Features \cite{RagISBA2019} & 46.82 & 86.62 & 81.64 & 12.32 & 19.72 & 13.2  & 9.74  & 15.95 & 8.91 \bigstrut\\
\cline{2-12}          & \multicolumn{1}{c|}{\multirow{2}[4]{*}{MIPGAN-II}} & Ensemble Features \cite{EnsembleFeatures_2020} & 34.13 & 70.49 & 61.57 & 5.32  & 6.68  & 2.57  & 6.72  & 8.16  & 4.14 \bigstrut\\
\cline{3-12}          &       & Hybrid Features \cite{RagISBA2019} & 44.96 & 83.7  & 75.47 & 5.9   & 8.42  & 3.23  & 5.67  & 6.18  & 2.91 \bigstrut \\
    \hline
    \end{tabular}%
    }
  \label{tab:MADlandmark1}%
\end{table*}%


\begin{table*}[htbp]
  \centering
  \caption{Quantitative performance of MAD - Training- Landmarks-II \cite{UBO_Morphing_Tool}}
  \resizebox{1\linewidth}{!}{
    \begin{tabular}{|c|c|c|c|c|c|c|c|c|c|c|c|}
    \hline
    \multicolumn{1}{|c|}{\multirow{3}[6]{*}{\textbf{Morph Generation Type: Training}}} & \multicolumn{1}{c|}{\multirow{3}[6]{*}{\textbf{Morph Generation Type: Testing}}} & \multirow{3}[6]{*}{\textbf{MAD Algorithms}} & \multicolumn{3}{c|}{\textbf{Digital}} & \multicolumn{3}{c|}{\textbf{Print-scan}} & \multicolumn{3}{c|}{\textbf{Print-scan with compression}} \bigstrut\\
\cline{4-12}          &       & \multicolumn{1}{c|}{} & \multicolumn{1}{c|}{\multirow{2}[4]{*}{D-EER(\%)}} & \multicolumn{2}{c|}{BPCER @ APCER =  } & \multicolumn{1}{c|}{\multirow{2}[4]{*}{D-EER(\%)}} & \multicolumn{2}{c|}{BPCER @ APCER =  } & \multicolumn{1}{c|}{\multirow{2}[4]{*}{D-EER(\%)}} & \multicolumn{2}{c|}{BPCER @ APCER =  } \bigstrut\\
\cline{5-6}\cline{8-9}\cline{11-12}          &       & \multicolumn{1}{r|}{} &       & 5\%   & 10\%  &       & 5\%   & 10\%  &       & 5\%   & 10\% \bigstrut\\
    \hline
    \multicolumn{1}{|c|}{\multirow{10}[20]{*}{\textbf{Landmarks-II \cite{UBO_Morphing_Tool}}}} & \multicolumn{1}{c|}{\multirow{2}[4]{*}{Landmarks-I \cite{raghavendra2017face}}} & Ensemble Features \cite{EnsembleFeatures_2020} & 48.57 & 97.77 & 95.36 & 24.19 & 52.48 & 43.22 & 21.64 & 47.51 & 36.19 \bigstrut\\
\cline{3-12}          &       & Hybrid Features \cite{RagISBA2019} & 45.67 & 96.91 & 94.16 & 32.26 & 77.87 & 66.55 & 24.51 & 50.94 & 40.65 \bigstrut\\
\cline{2-12}          & \cellcolor{mygray} & \cellcolor{mygray}Ensemble Features \cite{EnsembleFeatures_2020} & \cellcolor{mygray}3.62  & \cellcolor{mygray}2.22  & \cellcolor{mygray}0.68  & \cellcolor{mygray}6.32  & \cellcolor{mygray}7.97  & \cellcolor{mygray}2.42  & \cellcolor{mygray}5.57  & \cellcolor{mygray}6.41  & \cellcolor{mygray}2.42 \bigstrut\\
\cline{3-12}          &  \multicolumn{1}{c|}{\multirow{-2}[0]{*}{\cellcolor{mygray}Landmarks-II \cite{UBO_Morphing_Tool}}}     & \cellcolor{mygray}Hybrid Features \cite{RagISBA2019} & \cellcolor{mygray}1.53  & \cellcolor{mygray}0.17  & \cellcolor{mygray}0     & \cellcolor{mygray}5.21  & \cellcolor{mygray}5.19  & \cellcolor{mygray}3.14  & \cellcolor{mygray}5.37  & \cellcolor{mygray}5.71  & \cellcolor{mygray}3.46 \bigstrut\\
\cline{2-12}          & \multicolumn{1}{c|}{\multirow{2}[4]{*}{StyleGAN \cite{MorphStyleGAN2020}}} & Ensemble Features \cite{EnsembleFeatures_2020} & 29.67 & 61.92 & 52.48 & 27.18 & 61.57 & 50.6  & 29.18 & 62.14 & 52.48 \bigstrut\\
\cline{3-12}          &       & Hybrid Features \cite{RagISBA2019} & 34.76 & 74.44 & 62.95 & 34.8  & 67.23 & 58.14 & 23.17 & 49.22 & 38.25 \bigstrut\\
\cline{2-12}          & \multicolumn{1}{c|}{\multirow{2}[4]{*}{MIPGAN-I}} & Ensemble Features \cite{EnsembleFeatures_2020} & 30.23 & 65.35 & 53.17 & 43.92 & 87.65 & 79.24 & 44.24 & 89.23 & 82.33 \bigstrut\\
\cline{3-12}          &       & Hybrid Features \cite{RagISBA2019} & 46.29 & 84.04 & 77.01 & 34.16 & 71.18 & 64.66 & 35.5  & 76.84 & 65.52 \bigstrut\\
\cline{2-12}          & \multicolumn{1}{c|}{\multirow{2}[4]{*}{MIPGAN-II}} & Ensemble Features \cite{EnsembleFeatures_2020} & 27.13 & 58.83 & 45.45 & 33.57 & 77.35 & 65.52 & 40.46 & 84.9  & 75.47 \bigstrut\\
\cline{3-12}          &       & Hybrid Features \cite{RagISBA2019} & 46.82 & 83.53 & 75.81 & 35.91 & 77.18 & 65.24 & 36.5  & 79.24 & 68.78 \bigstrut\\
    \hline
    \end{tabular}%
    }
  \label{tab:MADlandmark2}%
\end{table*}%

\begin{table*}[htbp]
  \centering
  \caption{Quantitative performance of MAD - Training- StyleGAN \cite{MorphStyleGAN2020}}
  \resizebox{1\linewidth}{!}{
    \begin{tabular}{|c|c|c|c|c|c|c|c|c|c|c|c|}
    \hline
    \multicolumn{1}{|c|}{\multirow{3}[6]{*}{\textbf{Morph Generation Type: Training}}} & \multicolumn{1}{c|}{\multirow{3}[6]{*}{\textbf{Morph Generation Type: Testing}}} & \multirow{3}[6]{*}{\textbf{MAD Algorithms}} & \multicolumn{3}{c|}{\textbf{Digital}} & \multicolumn{3}{c|}{\textbf{Print-scan}} & \multicolumn{3}{c|}{\textbf{Print-scan with compression}} \bigstrut\\
\cline{4-12}          &       & \multicolumn{1}{c|}{} & \multicolumn{1}{c|}{\multirow{2}[4]{*}{D-EER(\%)}} & \multicolumn{2}{c|}{BPCER @ APCER =  } & \multicolumn{1}{c|}{\multirow{2}[4]{*}{D-EER(\%)}} & \multicolumn{2}{c|}{BPCER @ APCER =  } & \multicolumn{1}{c|}{\multirow{2}[4]{*}{D-EER(\%)}} & \multicolumn{2}{c|}{BPCER @ APCER =  } \bigstrut\\
\cline{5-6}\cline{8-9}\cline{11-12}          &       & \multicolumn{1}{c|}{} &       & 5\%   & 10\%  &       & 5\%   & 10\%  &       & 5\%   & 10\% \bigstrut\\
    \hline
    \multicolumn{1}{|c|}{\multirow{10}[20]{*}{\textbf{StyleGAN \cite{MorphStyleGAN2020}}}} & \multicolumn{1}{c|}{\multirow{2}[4]{*}{Landmarks-I \cite{raghavendra2017face}}} & Ensemble Features \cite{EnsembleFeatures_2020} & 0.32  & 0     & 0     & 16.6  & 28.13 & 19.89 & 13.89 & 22.12 & 17.66 \bigstrut\\
\cline{3-12}          &       & Hybrid Features \cite{RagISBA2019} & 0.42  & 0     & 0     & 15.26 & 26.41 & 17.66 & 14.37 & 22.81 & 16.92 \bigstrut\\
\cline{2-12}          & \multicolumn{1}{c|}{\multirow{2}[4]{*}{Landmarks-II \cite{UBO_Morphing_Tool}}} & Ensemble Features \cite{EnsembleFeatures_2020} & 44.72 & 89.53 & 80.61 & 38.31 & 78.5  & 69.15 & 38.84 & 83.7  & 74.17 \bigstrut\\
\cline{3-12}          &       & Hybrid Features \cite{RagISBA2019} & 45.65 & 90.22 & 84.56 & 34.18 & 81.95 & 70.53 & 32.93 & 78.5  & 64.12 \bigstrut\\
\cline{2-12}          & \cellcolor{mygray} & \cellcolor{mygray}Ensemble Features \cite{EnsembleFeatures_2020} &\cellcolor{mygray} 0     &\cellcolor{mygray} 0     &\cellcolor{mygray} 0     & \cellcolor{mygray}0     &\cellcolor{mygray} 0     & \cellcolor{mygray}0     & \cellcolor{mygray}0     & \cellcolor{mygray}0     & \cellcolor{mygray}0 \bigstrut\\
\cline{3-12}          & \multicolumn{1}{c|}{\multirow{-2}[0]{*}{\cellcolor{mygray}StyleGAN \cite{MorphStyleGAN2020}}}      & \cellcolor{mygray}Hybrid Features \cite{RagISBA2019} & \cellcolor{mygray}0     & \cellcolor{mygray}0     & \cellcolor{mygray}0     & \cellcolor{mygray}0     &\cellcolor{mygray} 0     & \cellcolor{mygray}0     &\cellcolor{mygray} 0     &\cellcolor{mygray} 0     & \cellcolor{mygray}0 \bigstrut\\
\cline{2-12}          & \multicolumn{1}{c|}{\multirow{2}[4]{*}{MIPGAN-I}} & Ensemble Features \cite{EnsembleFeatures_2020} & 39.97 & 75.98 & 68.78 & 20.21 & 42.14 & 33.44 & 20.73 & 45.28 & 36.53 \bigstrut\\
\cline{3-12}          &       & Hybrid Features \cite{RagISBA2019} & 46.45 & 86.79 & 77.87 & 29.34 & 59.19 & 47.51 & 24.87 & 51.62 & 41.18 \bigstrut\\
\cline{2-12}          & \multicolumn{1}{c|}{\multirow{2}[4]{*}{MIPGAN-II}} & Ensemble Features \cite{EnsembleFeatures_2020} & 39.93 & 73.58 & 66.89 & 15.78 & 28.14 & 19.38 & 13.72 & 28.98 & 16.63 \bigstrut\\
\cline{3-12}          &       & Hybrid Features \cite{RagISBA2019} & 44.72 & 82.16 & 73.75 & 19.36 & 43.22 & 28.64 & 16.98 & 32.93 & 23.84 \bigstrut\\
    \hline
    \end{tabular}%
    }
  \label{tab:MADstylegan}%
\end{table*}%


\begin{table*}[htbp]
  \centering
  \caption{Quantitative performance of MAD - Training- MIPGAN-I}
  \resizebox{1\linewidth}{!}{
    \begin{tabular}{|c|c|c|c|c|c|c|c|c|c|c|c|}
    \hline
    \multicolumn{1}{|c|}{\multirow{3}[6]{*}{\textbf{Morph Generation Type: Training}}} & \multicolumn{1}{c|}{\multirow{3}[6]{*}{\textbf{Morph Generation Type: Testing}}} & \multirow{3}[6]{*}{\textbf{MAD Algorithms}} & \multicolumn{3}{c|}{\textbf{Digital}} & \multicolumn{3}{c|}{\textbf{Print-scan}} & \multicolumn{3}{c|}{\textbf{Print-scan with compression}} \bigstrut\\
\cline{4-12}          &       & \multicolumn{1}{c|}{} & \multicolumn{1}{c|}{\multirow{2}[4]{*}{D-EER(\%)}} & \multicolumn{2}{c|}{BPCER @ APCER =  } & \multicolumn{1}{c|}{\multirow{2}[4]{*}{D-EER(\%)}} & \multicolumn{2}{c|}{BPCER @ APCER =  } & \multicolumn{1}{c|}{\multirow{2}[4]{*}{D-EER(\%)}} & \multicolumn{2}{c|}{BPCER @ APCER =  } \bigstrut\\
\cline{5-6}\cline{8-9}\cline{11-12}          &       & \multicolumn{1}{c|}{} &       & 5\%   & 10\%  &       & 5\%   & 10\%  &       & 5\%   & 10\% \bigstrut\\
    \hline
    \multicolumn{1}{|c|}{\multirow{10}[20]{*}{\textbf{MIPGAN-I}}} & \multicolumn{1}{c|}{\multirow{2}[4]{*}{Landmarks-I \cite{raghavendra2017face}}} & Ensemble Features \cite{EnsembleFeatures_2020} & 23.66 & 51.45 & 39.96 & 5.82  & 7.22  & 2.92  & 6.17  & 7.54  & 3.94 \bigstrut\\
\cline{3-12}          &       & Hybrid Features \cite{RagISBA2019} & 47.15 & 87.16 & 79.41 & 6.5   & 8.23  & 4.15  & 7.91  & 10.29 & 6.34 \bigstrut\\
\cline{2-12}          & \multicolumn{1}{c|}{\multirow{2}[4]{*}{Landmarks-II \cite{UBO_Morphing_Tool}}} & Ensemble Features \cite{EnsembleFeatures_2020} & 35.38 & 82.33 & 68.95 & 41.67 & 95.14 & 83.53 & 43.68 & 96.01 & 85.44 \bigstrut\\
\cline{3-12}          &       & Hybrid Features \cite{RagISBA2019} & 28.62 & 75.64 & 61.4  & 44.38 & 95.66 & 85.78 & 38.18 & 90.46 & 78.16 \bigstrut\\
\cline{2-12}          & \multicolumn{1}{c|}{\multirow{2}[4]{*}{StyleGAN \cite{MorphStyleGAN2020}}} & Ensemble Features \cite{EnsembleFeatures_2020} & 17.72 & 37.22 & 26.58 & 12.19 & 26.24 & 15.26 & 11.82 & 24.69 & 14.23 \bigstrut\\
\cline{3-12}          &       & Hybrid Features \cite{RagISBA2019} & 31.16 & 64.32 & 53.85 & 11.99 & 19.2  & 13.72 & 9.93  & 18.15 & 9.94 \bigstrut\\
\cline{2-12}          & \cellcolor{mygray} & \cellcolor{mygray}Ensemble Features \cite{EnsembleFeatures_2020} & \cellcolor{mygray}0     & \cellcolor{mygray}0     & \cellcolor{mygray}0     & \cellcolor{mygray}0     & \cellcolor{mygray}0     & \cellcolor{mygray}0     & \cellcolor{mygray}0     & \cellcolor{mygray}0     & \cellcolor{mygray}0 \bigstrut\\
\cline{3-12}          & \multicolumn{1}{c|}{\cellcolor{mygray}\multirow{-2}[0]{*}{MIPGAN-I}}      & \cellcolor{mygray}Hybrid Features \cite{RagISBA2019} & \cellcolor{mygray}0     & \cellcolor{mygray}0     & \cellcolor{mygray}0     & \cellcolor{mygray}0     & \cellcolor{mygray}0     & \cellcolor{mygray}0     & \cellcolor{mygray}0     & \cellcolor{mygray}0     & \cellcolor{mygray}0 \bigstrut\\
\cline{2-12}          & \multicolumn{1}{c|}{\multirow{2}[4]{*}{MIPGAN-II}} & Ensemble Features \cite{EnsembleFeatures_2020} & 2.15  & 0.17  & 0     & 0.68  & 0     & 0     & 0.64  & 0     & 0 \bigstrut\\
\cline{3-12}          &       & Hybrid Features \cite{RagISBA2019} & 1.36  & 0.34  & 0     & 0.86  & 0     & 0     & 0.8461 & 0     & 0 \bigstrut\\
    \hline
    \end{tabular}%
    }
  \label{tab:MADMIPGAN1}%
\end{table*}%


\begin{table*}[htbp]
  \centering
  \caption{Quantitative performance of MAD - Training- MIPGAN-II}
  \resizebox{1\linewidth}{!}{
    \begin{tabular}{|c|c|p{10.435em}|c|c|c|c|c|c|c|c|c|}
    \hline
    \multicolumn{1}{|r|}{\multirow{3}[6]{*}{\textbf{Morph Generation Type: Training}}} & \multicolumn{1}{r|}{\multirow{3}[6]{*}{\textbf{Morph Generation Type: Testing}}} & \multirow{3}[6]{*}{\textbf{MAD Algorithms}} & \multicolumn{3}{p{12.705em}|}{\textbf{Digital}} & \multicolumn{3}{p{12.705em}|}{\textbf{Print-scan}} & \multicolumn{3}{p{12.705em}|}{\textbf{Print-scan with compression}} \bigstrut\\
\cline{4-12}          &       & \multicolumn{1}{r|}{} & \multicolumn{1}{c|}{\multirow{2}[4]{*}{D-EER(\%)}} & \multicolumn{2}{p{8.6em}|}{BPCER @ APCER =  } & \multicolumn{1}{c|}{\multirow{2}[4]{*}{D-EER(\%)}} & \multicolumn{2}{p{8.6em}|}{BPCER @ APCER =  } & \multicolumn{1}{c|}{\multirow{2}[4]{*}{D-EER(\%)}} & \multicolumn{2}{p{8.6em}|}{BPCER @ APCER =  } \bigstrut\\
\cline{5-6}\cline{8-9}\cline{11-12}          &       & \multicolumn{1}{r|}{} &       & 5\%   & 10\%  &       & 5\%   & 10\%  &       & 5\%   & 10\% \bigstrut\\
    \hline
    \multicolumn{1}{|c|}{\multirow{10}[20]{*}{\textbf{MIPGAN-II}}} & \multicolumn{1}{c|}{\multirow{2}[4]{*}{Landmarks-I \cite{raghavendra2017face}}} & Ensemble Features \cite{EnsembleFeatures_2020} & 13.08 & 29.15 & 15.78 & 4.28  & 3.94  & 2.22  & 4.28  & 3.61  & 2.22 \bigstrut\\
\cline{3-12}          &       & Hybrid Features \cite{RagISBA2019} & 40.14 & 77.7  & 67.23 & 5.49  & 5.48  & 2.4   & 7.21  & 10.98 & 4.15 \bigstrut\\
\cline{2-12}          & \multicolumn{1}{c|}{\multirow{2}[4]{*}{Landmarks-II \cite{UBO_Morphing_Tool}}} & Ensemble Features \cite{EnsembleFeatures_2020} & 32.37 & 84.9  & 70.32 & 39.2  & 90.12 & 82.32 & 44.17 & 95.49 & 88.73 \bigstrut\\
\cline{3-12}          &       & Hybrid Features \cite{RagISBA2019} & 23.88 & 63.8  & 45.62 & 40.22 & 88.9  & 79.2  & 38.96 & 94.28 & 82.14 \bigstrut\\
\cline{2-12}          & \multicolumn{1}{c|}{\multirow{2}[4]{*}{StyleGAN \cite{MorphStyleGAN2020}}} & Ensemble Features \cite{EnsembleFeatures_2020} & 12.51 & 22.29 & 15.78 & 13.72 & 29.67 & 18.18 & 14.25 & 31.73 & 20.41 \bigstrut\\
\cline{3-12}          &       & Hybrid Features \cite{RagISBA2019} & 24.7  & 49.74 & 41.85 & 12.87 & 26.58 & 14.75 & 11.86 & 26.92 & 15.09 \bigstrut\\
\cline{2-12}          & \multicolumn{1}{c|}{\multirow{2}[4]{*}{MIPGAN-I}} & Ensemble Features \cite{EnsembleFeatures_2020} & 1.56  & 0.68  & 0.34  & 2.14  & 1.22  & 0.53  & 2.57  & 0.85  & 0.34 \bigstrut\\
\cline{3-12}          &       & Hybrid Features \cite{RagISBA2019} & 2.27  & 0.85  & 0.17  & 4.79  & 4.8   & 3.43  & 4.3   & 3.6   & 2.22 \bigstrut\\
\cline{2-12}          & \cellcolor{mygray} & \cellcolor{mygray}Ensemble Features \cite{EnsembleFeatures_2020} & \cellcolor{mygray}0     & \cellcolor{mygray}0     & \cellcolor{mygray}0     & \cellcolor{mygray}0     & \cellcolor{mygray}0     & \cellcolor{mygray}0     & \cellcolor{mygray}0     & \cellcolor{mygray}0     & \cellcolor{mygray}0 \bigstrut\\
\cline{3-12}          &  \multicolumn{1}{c|}{\cellcolor{mygray}\multirow{-2}[0]{*}{MIPGAN-II}}     & \cellcolor{mygray}Hybrid Features \cite{RagISBA2019} & \cellcolor{mygray}0     & \cellcolor{mygray}0     & \cellcolor{mygray}0     & \cellcolor{mygray}0     & \cellcolor{mygray}0     & \cellcolor{mygray}0     & \cellcolor{mygray}0     & \cellcolor{mygray}0     & \cellcolor{mygray}0 \bigstrut\\
    \hline
    \end{tabular}%
    }
  \label{tab:MADMIPGAN2}%
\end{table*}%

Considering the success rate of the newly generated dataset, we naturally choose to evaluate the morphing attack detection performance to also validate the robustness of  existing MAD mechanisms. Additionally, we investigate recent works about general face manipulation detection \cite{rossler2019faceforensics++} \cite{jain2020detecting} \cite{PartialFace_2019} and some results are shown in the supplementary material. In this work, we focus on single image based morphing attack detection (S-MAD) as it perfectly suits our dataset.  MAD has been widely addressed in the literature by developing the techniques based on both deep learning \cite{Matteo_PrintScan_2019}, \cite{StyleYourMorph_Ceibold_2019},  \cite{Dempster_Shafer_2019} \cite{BorderCtrl_MAD_2020} \cite{Peng_2019} and non-deep learning \cite{StirTrace_Hildebrandt_2017} \cite{IWBF2017_StirTrace} \cite{seibold2018reflection} \cite{Kraetzer:2017:MAP:3082031.3083244} approaches. Readers can refer to \cite{venkatesh2020facemorphingSurvey} for an exclusive survey on face MAD.  Owing to the recent works detailing the applicability of Hybrid features  \cite{RagISBA2019} and Ensemble features \cite{EnsembleFeatures_2020} in detecting morphing attacks, we choose to benchmark both  Hybrid features \cite{RagISBA2019} and Ensemble features \cite{EnsembleFeatures_2020}. While the Hybrid features \cite{RagISBA2019} resort to extracting features using both scale space and color space combined with multiple classifiers, Ensemble features \cite{EnsembleFeatures_2020} employ a variety of textural features in conjunction with a set of classifiers. In common both approaches evaluate a wide variety of MAD mechanisms in a holistic manner supported by empirical results \cite{RagISBA2019,  EnsembleFeatures_2020}. In addition, the Hybrid features \cite{RagISBA2019} mechanisms are also validated against the ongoing NIST FRVT MORPH challenge \cite{NistFrvtMorph} with the best performance in detecting printed and scanned morph images justifying our selection of algorithm to benchmark the newly composed database. 

\begin{figure*}[htp]
	\centering
	\includegraphics[width=0.7\linewidth]{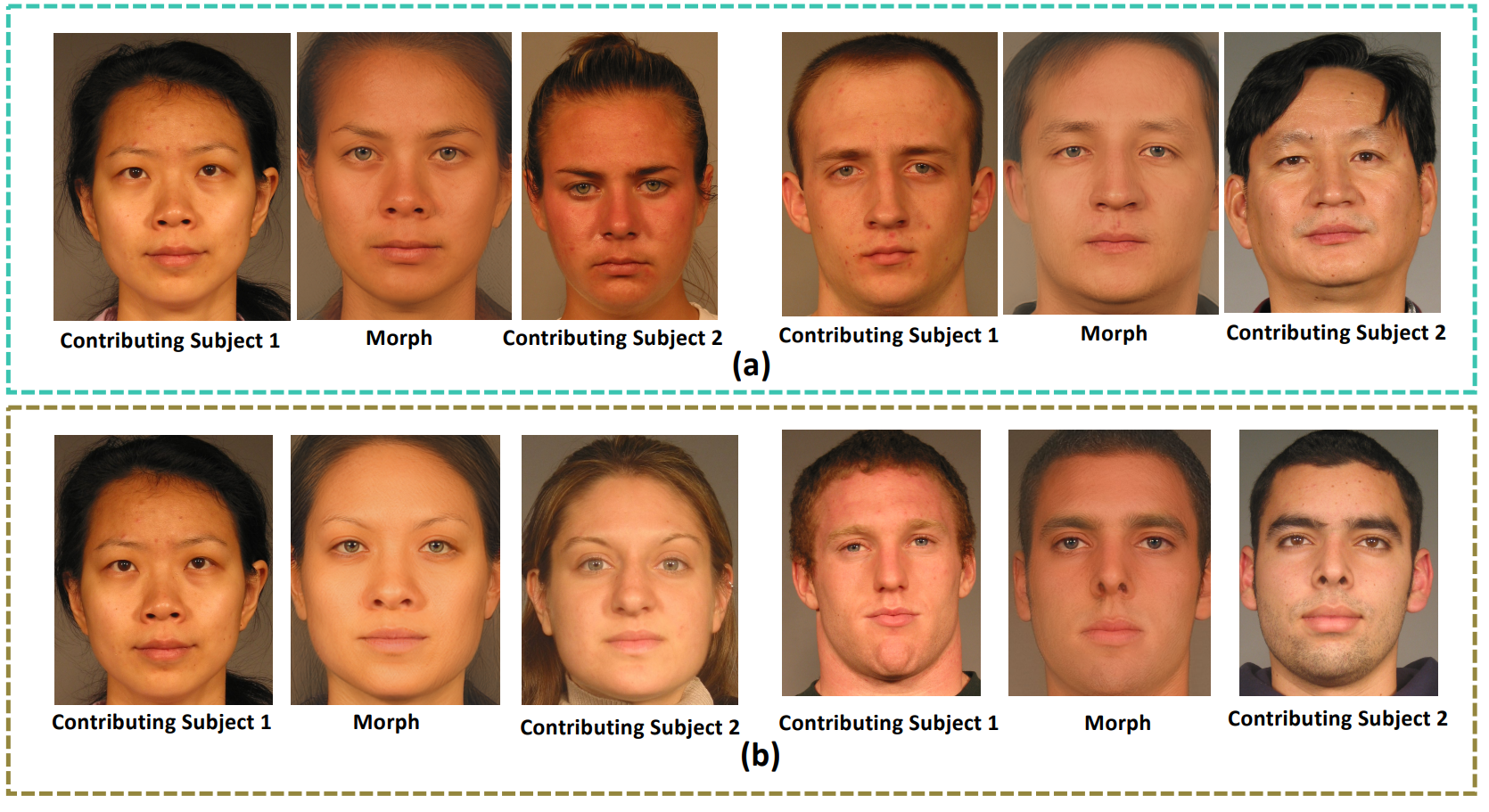}
	\caption{Examples of  morphed images that failed to attack FRS (a) morphed face images generated using proposed MIPGAN-I (b) morphed face images generated using proposed MIPGAN-II}
	\label{fig:failVerifiy}
\end{figure*}

The reporting of MAD performance is following the ISO/IEC metrics \cite{ISO-IEC-30107-3-PAD-metrics-170227} namely the  Attack Presentation Classification Error Rate (APCER ($\%$)) which defines the proportion of attack images (morph images) incorrectly classified as bona fide images and the Bona fide Presentation Classification Error Rate (BPCER ($\%$)) in which bona fide images incorrectly classified as attack images are counted \cite{ISO-IEC-30107-3-PAD-metrics-170227}  along with the Detection Equal Error Rate (D-EER ($\%$)). To evaluate the generated morphed face image's attack potential, we have sub-divided the newly generated database into two sets for training and testing that consists of independent data subjects with no overlap between the splits. The training set includes 690 bona fide images and 1190 morphed images. The testing set consists of 580 bona fide and 1310 morphed images. To effectively evaluate the performance of the MAD reflecting a real-life scenario, we report the results on both intra (training and testing dataset from the same morph generation approach) and inter (training on one type of morphing techniques and testing on another type of morphing techniques) evaluation of MAD mechanisms. Extensive experiments are performed on digital, print-scan and print-scan with compression data types to provide an in-depth analysis of the S-MAD performance. 
Table ~\ref{tab:MADlandmark1}, ~\ref{tab:MADlandmark2}, ~\ref{tab:MADstylegan}, ~\ref{tab:MADMIPGAN1} and ~\ref{tab:MADMIPGAN2} presents the quantitative results of MAD mechanisms on morph generation methods together with the  SOTA morph generation techniques. Based on the results obtained from the intra-dataset experiments, we make some concrete observations as listed below:
\begin{itemize}
\item  The intra-dataset evaluation indicates that the morphing attacks are detected with a good success rate irrespective of the type of generation. 
\item In general, the attack detection success rate is high with digital data when compared to print-scan and print-scan compression. 
\item Among the different types of morph generation techniques, the Landmark-II based morph generation shows the highest error rates. The attack images created using StyleGAN and proposed MIPGAN can be efficiently detected using both the employed approaches with high accuracy. This can be attributed to the noises that are synthesised using GANs due to the computational modifications performed on the latent space in GAN-based morph generation methods. 
\end{itemize}

In the following, we discuss the important observations based on the results obtained from inter-dataset MAD analysis:  

\begin{itemize}
\item  The performance of the MAD techniques are degraded on all five different case studies as indicated in the Table ~\ref{tab:MADlandmark1}, ~\ref{tab:MADlandmark2}, ~\ref{tab:MADstylegan}, ~\ref{tab:MADMIPGAN1} and ~\ref{tab:MADMIPGAN2}. 

\item Training MAD algorithms with one type of landmarks-based method did not show the improvement in detection performance of another kind of landmarks-based morph generation method. 
\item When MAD mechanisms are trained using the Landmarks-I \cite{raghavendra2017face} method, the degraded performance is noted for all other morph generation methods except for the StyleGAN \cite{MorphStyleGAN2020} based approach. This fact is also noted when we train the MAD techniques using StyleGAN \cite{MorphStyleGAN2020} generated samples and test it with Landmarks-I \cite{raghavendra2017face} samples. Thus, the StyleGAN \cite{MorphStyleGAN2020} based morph generation is easy to detect even when MAD mechanisms are not trained using the images from same morph generation scheme.
\item When MAD algorithms are trained using Landmarks-II \cite{UBO_Morphing_Tool} samples, MAD algorithms indicate degraded performance on all other morph generation techniques. 
\item When MAD mechanisms are trained using the proposed MIPGAN-I generated samples. The MAD mechanisms indicate an excellent detection performance on MIPGAN-II samples. However, the detection performance of MAD methods is deceived with other morph generation techniques. 
\item It is interesting to note that when MAD mechanisms are trained using MIPGAN-I/MIPGAN-II, higher detection accuracy can be observed for print-scan and print-scan with compression data when compared to digital morph data. A possible reason is that the noise generated together with the morphed images using the proposed MIPGAN-I/MIPGAN-II can approximate the generated noise resulting from the print-scan and print-scan compression process. 
\item Based on the results of the inter-database MAD analysis, the detection of Landmarks-II \cite{UBO_Morphing_Tool} samples are challenging.
\end{itemize}

\section{Limitations of Current Work and Potential Future Works}
\label{sec:limitations-future-works}
Despite this work presenting a new approach to generate strong morphing attacks, which are empirically evaluated using COTS FRS, our work has a few noted limitations. In the current scope of work, we evaluate the impact of print and scan (re-digitizing) using one printer reflecting a realistic scenario. The MAD mechanism employed in this work has not been investigated with a wide range of printers and scanners that may impact the MAD performance. While we assert that the MAD performance may not vary extremely, when tested with a wider combination of printers and scanners, that empirical evaluation is yet to be conducted in future works.

A second aspect is that the proposed approach needs pre-selection of ethnicity for generating stronger attacks. Figure \ref{fig:failVerifiy} shows example morphed face images generated using the proposed method using MIPGAN-I and MIPGAN-II that fail to get verified to contributing subjects when ethnicity pre-selection is not performed \cite{raghavendra2017face}. We notice that the selection of contributing subjects plays an important role with the proposed method to generate stronger attacks with MIPGAN. It is our assertion that the selection of contributing subjects with similar geometric structures (particularly ethnicity and age) can improve the performance of the proposed system, but that aspect needs further investigation. 



\section{Conclusion}
\label{sec:conclusion}
Addressing the limitations of generating the strong and severe morphing attacks using GAN, we have proposed a new architecture for generating face morphed images in this work. The proposed approach (MIPGAN with two variants) for devising strong morphing attacks uses identity prior driven GAN with a customized loss exploiting perceptual quality and identity factors to generate realistic images that can strongly threaten FRS. In order to validate the attack potential of the proposed morph generation method, we have created a new dataset consisting of $30,000$  morphed images and $15,240$ bona fide images. Both COTS and deep learning based FRS were evaluated empirically to measure the success rate of the new approach and vulnerability was reported indicating the applicability of the new approach and newly generated database. In a similar direction, the dataset is also validated for detection performance by studying two state-of-art MAD mechanisms. Despite the high attack detection success rate by employed MAD, we note that the morphed images generated by MIPGAN can severely threaten FRS in a present state without MAD in FRS.

\balance
{\small
\bibliographystyle{IEEEtran}
	\bibliography{sushma-IWBF-2020-StyleGAN-200216}
}

\end{document}